%% file: arxiv.tex
\documentclass{article}

\usepackage{microtype}
\usepackage{graphicx}
\usepackage{subcaption}
\usepackage{booktabs} 
\usepackage{multirow}
\usepackage{enumitem}
\usepackage[dvipsnames]{xcolor}
\usepackage{xspace}
\usepackage{mathtools}
\usepackage{amssymb}
\usepackage{stfloats} \usepackage[rightcaption]{sidecap} 
\usepackage[utf8]{inputenc} \usepackage[T1]{fontenc}  \usepackage{amsfonts}       \usepackage{nicefrac}       \usepackage{url}            \usepackage{overpic}
\usepackage{colortbl}
\usepackage[full]{textcomp}
\usepackage{algorithm}
\usepackage{algorithmic}
\usepackage{wrapfig}

\usepackage[multidot]{grffile} 

\renewcommand{\textrightarrow}{$\rightarrow$}

\newcommand{\RR}{\mathbb{R}}

\newcommand{\sP}{\mathcal{P}}

\newcommand{\cscore}{{C-score}\xspace}
\newcommand{\cscores}{{C-scores}\xspace}
\newcommand{\arroweg}[2]{{\textsc{#1}\textrightarrow \textsc{#2}}}
\newcommand{\suplof}{\text{\scalebox{.5}[1.0]{LOF}}}

\definecolor{figorange}{RGB}{237,125,49}
\definecolor{figblue}{RGB}{91,155,213}

\usepackage{hyperref}

\usepackage[accepted]{icml2021}

\icmltitlerunning{Characterizing Structural Regularities of Labeled Data in Overparameterized Models}

\begin{document}

\twocolumn[
\icmltitle{Characterizing Structural Regularities of Labeled Data\\ in Overparameterized Models}

\icmlsetsymbol{equal}{*}

\begin{icmlauthorlist}
\icmlauthor{Ziheng Jiang}{equal,uw,octo,goow}
\icmlauthor{Chiyuan Zhang}{equal,goo}
\icmlauthor{Kunal Talwar}{goo,apple}
\icmlauthor{Michael C. Mozer}{goo,uc}
\end{icmlauthorlist}

\icmlaffiliation{goo}{Google Research, Brain Team, Mountain View, CA, USA.}
\icmlaffiliation{goow}{Work done while interning at Google.}
\icmlaffiliation{uw}{Paul G. Allen School of Computer Science, University of Washington, Seattle, WA, USA.}
\icmlaffiliation{octo}{OctoML.ai, Seattle, WA, USA.}
\icmlaffiliation{apple}{Presently at Apple Inc., Cupertino, CA, USA.}
\icmlaffiliation{uc}{Department of Computer Science, University of Colorado Boulder, Boulder, CO, USA.}

\icmlcorrespondingauthor{Chiyuan Zhang}{chiyuan@google.com}

\icmlkeywords{Machine Learning, ICML}

\vskip 0.3in
]

\printAffiliationsAndNotice{\icmlEqualContribution} 
\begin{abstract}
Humans are accustomed to environments that contain both regularities and exceptions. For example, at most gas stations, one pays prior to pumping, but the occasional rural station does not accept payment in advance.
Likewise, deep neural networks can generalize across instances that share common patterns or structures, yet have the capacity to memorize rare or irregular forms. We analyze how individual instances are treated by a model via a \emph{consistency score}. The score characterizes the expected accuracy for a held-out instance given training sets of varying size sampled from the data distribution. We obtain empirical estimates of this score for individual instances in multiple data sets, and we show that the score identifies out-of-distribution and mislabeled examples at one end of the continuum and strongly regular examples at the other end. We identify computationally inexpensive proxies to the consistency score using statistics collected during training. We show examples of potential applications to the analysis of deep-learning systems.
\end{abstract}

\section{Introduction}

Human learning requires both inferring regular patterns that generalize across many distinct examples and memorizing irregular examples.
The boundary between regular and irregular examples can be fuzzy.
For example, in learning
the past tense form of English verbs, there are some verbs whose past tenses must simply
be memorized (\arroweg{go}{went}, \arroweg{eat}{ate}, \arroweg{hit}{hit})
and there are many \emph{regular} verbs that obey the rule of appending ``ed''
(\arroweg{kiss}{kissed}, \arroweg{kick}{kicked}, \arroweg{brew}{brewed}, etc.).
Generalization to a novel word typically follows the ``ed'' rule, for example, \arroweg{bink}{binked}.
Intermediate between  the exception verbs and regular verbs are subregularities---a set
of exception verbs that have consistent structure
(e.g., the mapping of \arroweg{sing}{sang}, \arroweg{ring}{rang}).
Note that rule-governed and exception cases can have very similar forms, which increases
the difficulty of learning each. 
Consider one-syllable verbs containing `ee', which include the regular cases \arroweg{need}{needed}
as well as exception cases like \arroweg{seek}{sought}.
Generalization from the rule-governed cases can hamper the learning of the exception cases and vice-versa.
For instance, children in an environment where English is spoken
over-regularize by mapping \arroweg{go}{goed} early in the course of language learning.
Neural nets show the same interesting pattern for verbs over the course of training \citep{rumelhart1986}.

Intuitively, memorizing irregular examples is tantamount to building a look-up table with the individual facts accessible
for retrieval. Generalization requires the inference of statistical regularities in
the training environment, and the application of procedures or rules for exploiting
the regularities.
In deep learning, memorization is often considered a failure of a network because
memorization implies no generalization. However, mastering a domain involves knowing
when to generalize and when not to generalize, because the data manifolds are rarely unimodal. 

\begin{figure*}[t!]
    \centering    \includegraphics[width=\linewidth]{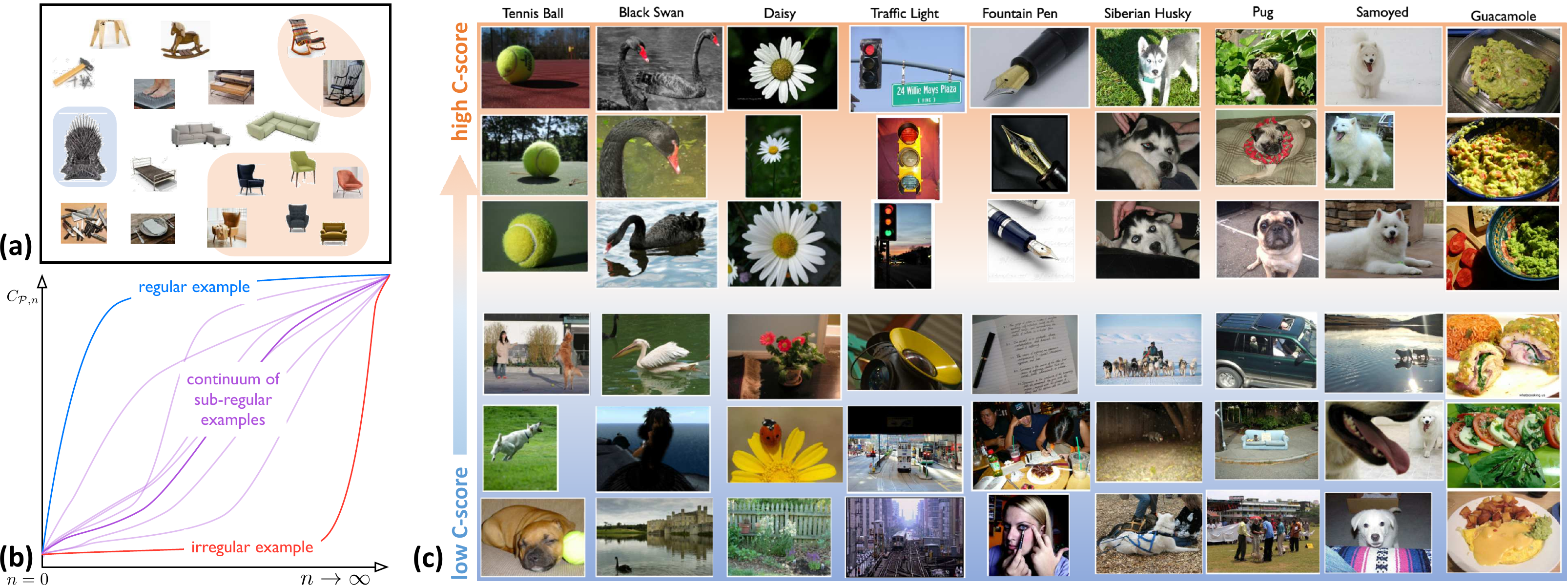}
    \caption{\small Regularities and exceptions in a binary chairs vs non-chairs problem. (b) illustration of consistency profiles.
    (c) Regularities  (\color{figorange}high \color{black} \cscores) and exceptions (\color{figblue}low \color{black} \cscores)
    in ImageNet.}
    \label{fig:chairs}
\end{figure*}

Consider the two-class problem of chair vs non-chair with training examples illustrated in Figure~\ref{fig:chairs}a.
The iron throne (lower left) forms a sparsely populated mode (\emph{sparse mode} for short) as there may not exist many similar cases in the data environment. 
Generic chairs (lower right) lie in a region with a consistent labeling (a densely populated mode, or \emph{dense mode}) and thus seems to follow a strong regularity. But there are many other cases in the continuum of the two extreme. For example, the rocking chair (upper right) has a few supporting neighbors but it lies in a distinct neighborhood from the majority of same-label instances (the generic chairs).

In this article, we formalize this continuum of the structural regularities of data sets in the context of training overparameterized deep networks. Let $D  \stackrel{n}{\sim} \sP$ be an i.i.d. sample of size $n$ from the underlying data distribution $\sP$, and $f(\cdot\,;D)$ be a model trained on $D$. For an instance $x$ with label $y$, we trace out the following \emph{consistency profile} by increasing $n$:
\begin{equation}
    C_{\sP,n}(x,y)=\mathbb{E}_{D \stackrel{n}{\sim} {\mathcal{P}}}
    [\mathbb{P}(f(x;D\backslash\{(x,y)\})=y],
    \label{eq:c-profile}
\end{equation}
Note by taking expectation over $(x,y)$, this measures the \emph{generalization performance} with respect to the underlying distribution $\mathcal{P}$. In contrast to the average behavior, we focus on the \emph{per-instance generalization} here, as it helps to reveal the internal regularity structures of the data distribution.
This article focuses on multi-class classification problems, but the definition can be easily extended to other problems by 
replacing the 0-1 classification loss with another suitable loss function.

$C_{\sP,n}(x,y)$ also encodes our high-level intuition about the structural regularities of the training data during (human or machine) learning.
In particular, we can characterize the multimodal structure of an underlying data distribution by grouping examples in terms of a model's generalization profile for those examples. An $(x,y)$ with high per-instance generalization lies in a region on the data manifold that is well supported by other regular instances.

For $n=0$, the model makes predictions entirely based on its prior belief. As $n$ increases, the model collects more information about $\sP$ and makes better predictions. For an $(x,y)$ instance belonging to a dense mode (e.g., the generic chairs in Figure~\ref{fig:chairs}a), the model prediction is accurate even for small $n$ because even small samples have many class-consistent neighbors. The blue curve in the cartoon sketch of Figure~\ref{fig:chairs}b illustrates this profile. For instances belonging to sparse modes (e.g., the iron throne in Figure~\ref{fig:chairs}a), the prediction will be inaccurate for even large $n$, as the red curve illustrates. Most instances fill the continuum between these two extreme cases, as illustrated by the purple curves in Figure~\ref{fig:chairs}b. To obtain a \emph{total ordering} for all examples, we pool the consistency profile into a scalar \emph{consistency score}, or \emph{\cscore} by taking expectation over $n$. Figure~\ref{fig:chairs}c shows examples from the ImageNet data set ranked by estimated \cscores, using a methodology we shortly describe. The images show that on many ImageNet classes, there exist dense modes of center-cropped, close-up shot of the representative examples; and at the other end of the \cscore ranking, there exist sparse modes of highly ambiguous examples (in many cases, the object is barely seen or can only be inferred from the context in the picture). 

With strong ties to both theoretical notions of generalization and human intuition, the consistency profile is an important
tool for understanding the regularity and subregularity structures of training data sets and the learning dynamics of models trained on those data.
The \cscore based ranking also has many potential uses, such as detecting
out-of-distribution and mislabeled instances; balancing learning between dense and sparse modes to ensure fairness when learning with data from underrepresented groups; or even as a diagnostic used to determine training priority in a curriculum learning setting \citep{bengio2009,saxena2019}.
In this article, we focus on formulating and analyzing consistency profiles, and apply the \cscore to analyzing the structure of real world image data sets and the learning dynamics of different optimizers. We also study efficient proxies and further applications to outlier detection.

Our key contributions are as follows:
\begin{itemize}[nosep,align=left,leftmargin=*]
    \item We formulate and analyze a consistency score that takes inspiration from generalization theory
    and show that it matches our intuitions about statistical regularities in natural-image data sets.
    \item We estimate the \cscores with a series of approximations and apply the measure
    to analyze the structural regularities of the MNIST, CIFAR-10, CIFAR-100, and \mbox{ImageNet} training sets.
    \item We evaluate computationally efficient proxies for the \cscore. We demonstrate that proxies based on distances between instances of the same class in latent space, while intuitively sensible, are in practice quite sensitive to the underlying distance metric. In contrast,
        learning-speed based proxies correlate very well with the \cscore. This observation is non-trivial because learning speed is measured on training examples and the \cscore is defined for hold-out generalization.     \item We demonstrate potential application of the \cscore as a tool for quantitative analysis of data sets, learning dynamics, and diagnosing and improving
    deep learning.
        \item To facilitate future research, we have released the pre-computed \cscores at the \href{https://pluskid.github.io/structural-regularity/}{project website}. 
    Model checkpoints, code, and extra visualizations are available too.
\end{itemize}

\section{Related Work}

Analyzing the structure of data sets has been a central topic for many fields like Statistics,
Data Mining and Unsupervised Learning. In this paper, we focus on supervised learning and
the interplay between the regularity structure of data and overparameterized neural network
learners. This differentiates our work from classical analyses based on input or (unsupervised) latent
representations. The distinction is especially prominent in deep learning where a supervised
learner jointly learns the classifier and the representation that captures the semantic
information in the labels.

In the context of deep supervised learning, \citet{protypical} proposed measures for identifying \emph{prototypical} examples which could serve
as a proxy for the complete data set and still achieve good performance. These examples are not necessarily
the center of a dense neighborhood, which is what our high \cscore measures. Two prototype measures explored in
\citet{protypical}, \emph{model confidence} and the \emph{learning speed}, are also measures we examine.
Their \emph{holdout retraining} and \emph{ensemble agreement} metrics are conceptually similar to our \cscore estimation algorithm. However, their retraining is a two-stage procedure involving pre-training and fine-tuning; their ensemble agreement mixes architectures with heterogeneous capacities and ignores labels. 
\citet{feldman2019does} and \citet{fz2019} studied the positive effects of memorization on generalization by measuring
the influence of a training example on a test example, and identifying pairs with strong influences. To 
quantify memorization, they defined a memorization score for each $(x,y)$ in a training set 
as the drop in prediction accuracy on $x$ when $(x,y)$ is removed. A point evaluation of our consistency profile on a fixed data size $n$ resembles the second term of their score. Our empirical C-score estimation is based on the estimator proposed in \citet{fz2019}. A key difference is that we are
interested in the profile with increasing $n$, i.e. the sample complexity required to correctly predict
$(x,y)$.

We evaluate various cheap-to-compute \emph{proxies} for the \cscore and found that the learning speed has a strong correlation with the \cscore.
Learning speed has been previously studied in contexts quite different from our focus on generalization of individual examples. \citet{mangalam2019deep} show that examples learned first are those that could be learned by shallower nets. \citet{hardt2016train} present theoretical results showing that the generalization gap is small if SGD training completes in relatively few steps. \citet{toneva2018empirical} study forgetting (the complement of learning speed) and informally relate forgetting to examples being outliers or mislabeled. There is a large literature of criteria with no explicit ties to generalization as the C-score has, but provides a means of stratifying instances. For example, \citet{wu2018blockdrop} measure the difficulty of an example by the number of residual blocks in a ResNet needed for prediction.

\section{The Consistency Profile and the \cscore}

\begin{figure}
\includegraphics[width=\linewidth]{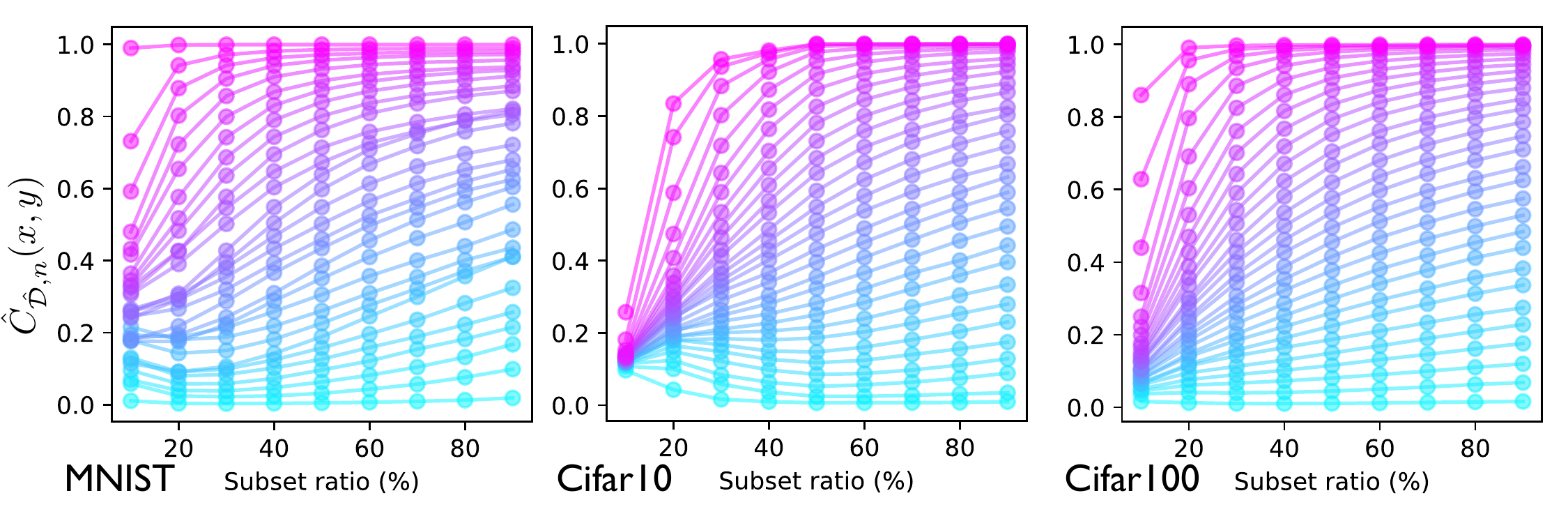}\vskip-10pt
\caption{\small Consistency profiles of training examples. Each curve in the figure corresponds to the average profile of a set of examples, partitioned according to the area under the profile curve of each example.}\label{fig:c-profile-3data}
\end{figure}

The consistency profile (Equation~\ref{eq:c-profile}) encodes the structural consistency of an example with the underlying data distribution $\sP$ via expected performance of models trained with increasingly large data sets sampled from $\sP$. However,
it is not possible to directly compute this profile because $\sP$ is generally unknown for typical learning problems. In practice, we usually have a fixed data set $\hat{\mathcal{D}}$ consisting of $N$ i.i.d. samples from $\sP$. So we can estimate the consistency profile with the following \emph{empirical consistency profile}:
\begin{equation}
    \hat{C}_{\hat{\mathcal{D}},n}(x,y) = 
    \hat{\mathbb{E}}^r_{D\stackrel{n}{\sim} \hat{\mathcal{D}}} \left[\mathbb{P}(f(x;D\backslash\{(x,y)\})=y)\right],
    \label{eq:empirical_cscore}
\end{equation}
where $n=0,1,\ldots,N-1$, $D$ is a subset of size $n$ uniformly sampled from $\hat{\mathcal{D}}$ excluding $(x,y)$, and $\hat{\mathbb{E}}^r$ denotes empirical averaging with $r$ i.i.d. samples of such subsets. To obtain a reasonably accurate estimate (say, $r=1000$), calculating the empirical consistency profile is still computationally prohibitive. For example, with
each of the 50,000 training example in the CIFAR-10 training set, we need
to train more than 2 trillion models. To obtain an estimate within the capability of current computation resources, we make two observations. First, model performance is generally stable when the training set size varies within a small range. Therefore, we can sample across the range of $n$ that we're concerned with and obtain the full profile via smooth interpolation. Second, let $D$ be a random subset of training data, then the single model $f(\cdot\,;D)$ can be reused in the estimation of all of the held-out examples $(x,y)\in\hat{\mathcal{D}}\backslash D$. As a result, with clever grouping and reuse, the number of models we need to train can be greatly reduced (See Algorithm~1 in the Appendix).

\begin{figure*}
\centering\includegraphics[width=.75\linewidth]{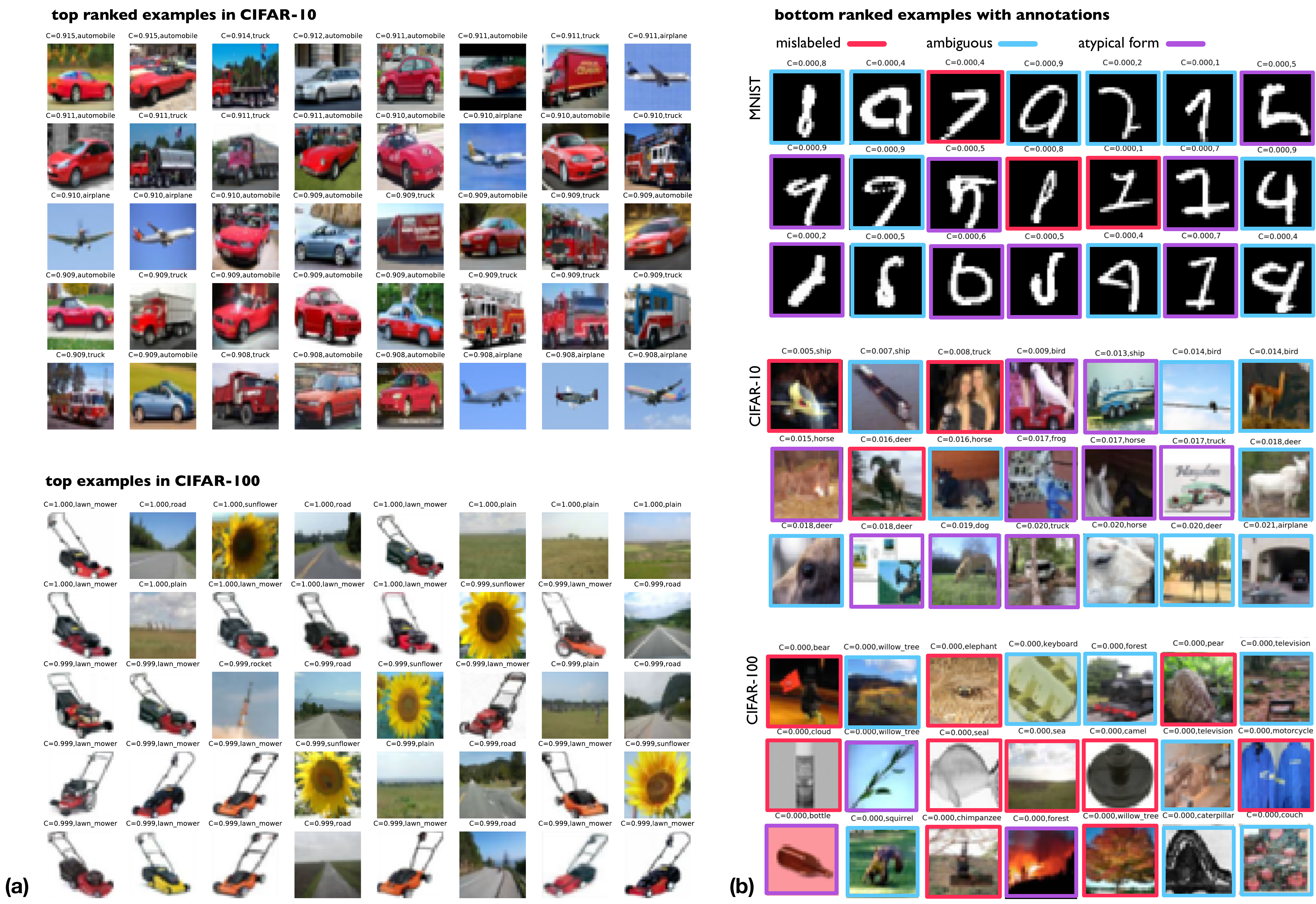}\vspace{-7pt}
\caption{\small(a) Top ranked examples in CIFAR-10 and CIFAR-100. (b) Bottom ranked examples with annotations.}
\label{fig:global-rank}
\end{figure*}

In particular, we sample $n$ dynamically according to the \emph{subset ratio} $s\in\{10\%,\ldots,90\%\}$ of the full available training set. We sample 2,000 subsets for the empirical expectation of each $n$ and visualize the estimated consistency profiles for clusters of similar examples in Figure~\ref{fig:c-profile-3data}. One interesting observation is that while CIFAR-100 is generally more difficult than CIFAR-10, the top ranked examples (magenta lines) in CIFAR-100 are more likely to be classified correctly when the subset ratio is low. 
Figure~\ref{fig:global-rank}a visualizes the top ranked examples from the two data sets. Note that in CIFAR-10, the dense modes from the truck and automobile classes are quite similar.

In contrast, Figure~\ref{fig:c-profile-3data} indicates that the bottom-ranked examples (cyan lines) have persistently
low probability of correct classification---sometimes below chance---even with a 90\% subset ratio.
We visualize some bottom-ranked examples and annotate them as (possibly) mislabeled, ambiguous (easily confused with another class or hard to identify the contents), and atypical form (e.g., burning ``forest'', fallen ``bottle''). As the subset ratio grows, regularities in the data distribution systematically pull the ambiguous instances in the wrong direction.
This behavior is analogous to the phenomenon we mentioned earlier that children over-regularize verbs
(\arroweg{go}{goed}) as they gain more linguistic exposure.

To get a total ordering of the examples in a data set, we distill the consistency profiles into a scalar \emph{consistency score}, or \cscore, by taking the expectation over $n$:
\begin{equation}
\hat{C}_{\hat{\mathcal{D}}}(x,y) = \mathbb{E}_n[\hat{C}_{\hat{\mathcal{D}},n}(x,y)]
\end{equation}
For the case where $n$ is sampled according to the subset ratio $s$, the expectation is taken over a uniform distribution over sampled subset sizes.

\section{The Structural Regularities of Common Image Data Sets}

We apply the \cscore estimate to analyze several common image data sets: MNIST~\citep{lecun1998gradient}, CIFAR-10 / CIFAR-100~\cite{cifar}, and ImageNet~\citep{russakovsky2015imagenet}. 
See the supplementary materials for details on architectures and hyperparameters.

\begin{figure*}
\centering
\includegraphics[width=.8\linewidth]{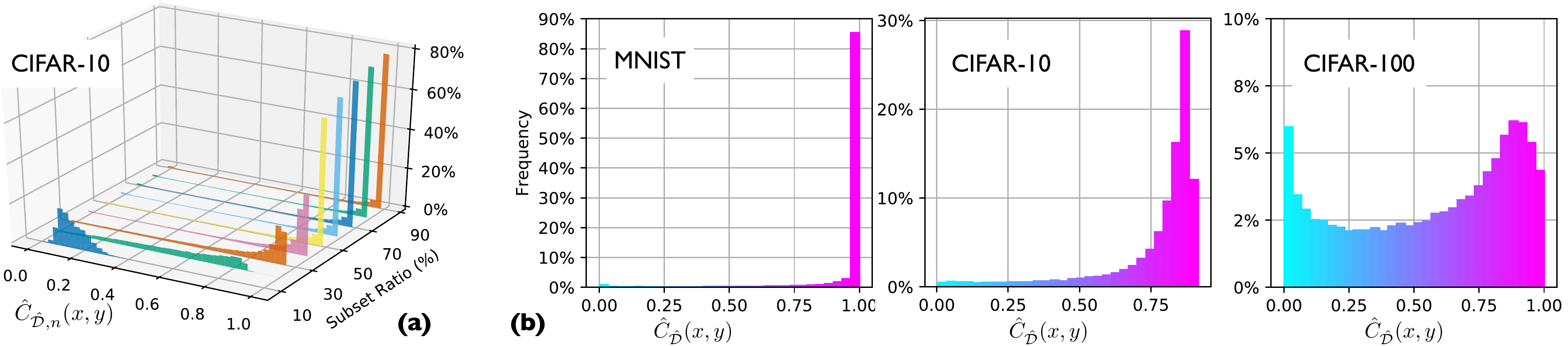}
\caption{(a) Histogram of $\hat{C}_{\hat{\mathcal{D}},n}$ for each subset ratio on CIFAR-10. (b) Histogram of the \cscore $\hat{C}_{\hat{\mathcal{D}}}$ averaged over all subset ratios on 3 different data sets.}
\label{fig:cscore-hist}
\end{figure*}

Figure~\ref{fig:cscore-hist}a shows the distribution of $\hat{C}_{\hat{\mathcal{D}},n}$ on CIFAR-10
for the values of $n$ corresponding to each subset ratio $s\in \{10,..., 90\}$. For each $s$, 2000 models are trained and held-out examples are evaluated. 
The Figure suggests that depending on $s$, instances may be concentrated near floor or ceiling, making them
difficult to distinguish (as we elaborate further shortly). 
By taking an expectation over $s$, the \cscore is less susceptible to floor and ceiling effects.
Figure~\ref{fig:cscore-hist}b shows
the histogram of this integrated \cscore on MNINT, CIFAR-10, and CIFAR-100.
The histogram of CIFAR-10 in Figure~\ref{fig:cscore-hist}b is distributed toward the high end, but is more uniformly spread than the histograms for specific subset ratios in Figure~\ref{fig:cscore-hist}a.

Visualization of examples ranked by the estimated score can be found in Figure~\ref{fig:global-rank}. Detailed per-class rankings can be found in the supplementary material.

Next we apply the \cscore analysis to the ImageNet data set. Training a standard model on ImageNet
costs one to two orders of magnitude more computing resources than training on CIFAR, preventing us from running
the \cscore estimation procedure described early. Instead, we investigated the feasibility of
approximating the \cscore with a \emph{point estimate}, i.e., selection of the $s$
that best represents the integral score. This is equivalent to taking expectation of $s$ with
respect to a point-mass distribution, as opposed to the uniform distribution over subset ratios.
By `best represents,' we mean that the ranking of
instances by the score matches the ranking by the score for a particular $s$.

\begin{figure}
\begin{overpic}[width=1.24in]{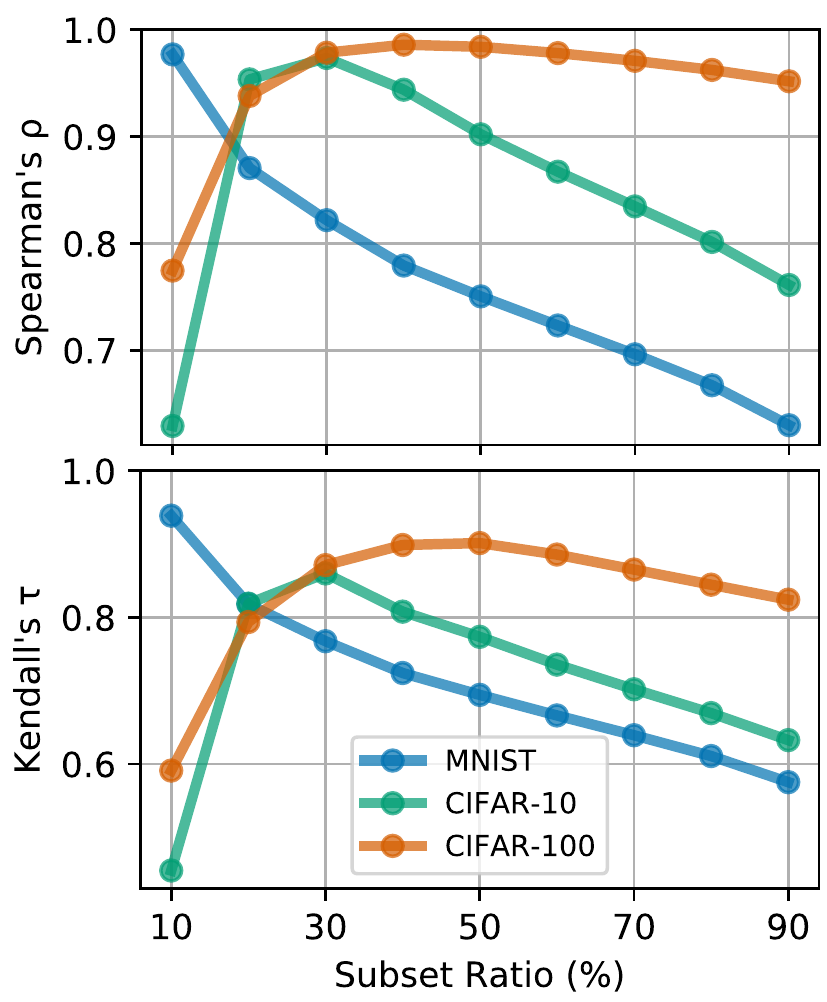}
\put(0,0){\footnotesize (a)}
\end{overpic}\hfill
\begin{overpic}[width=1.7in]{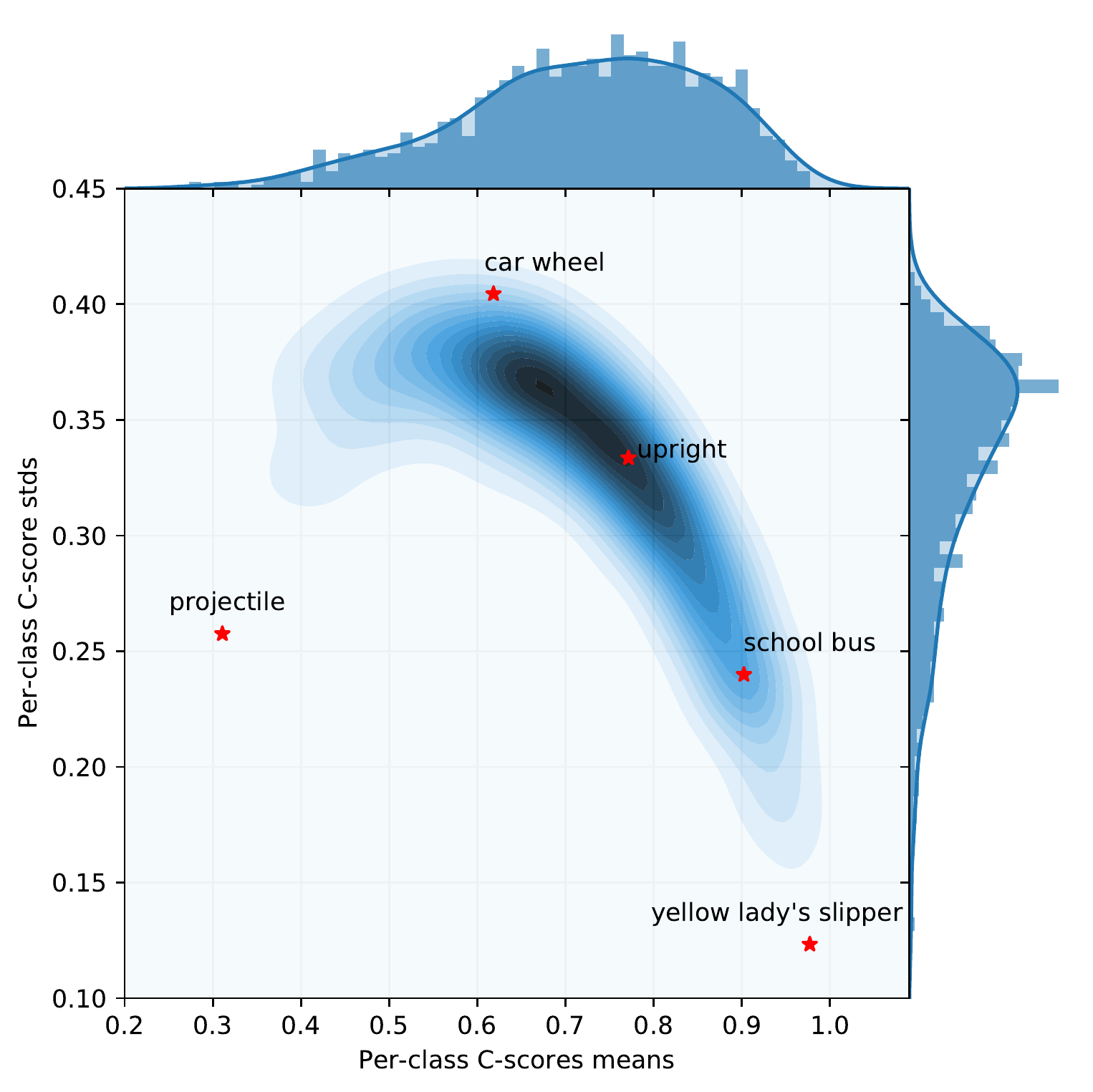}
\put(0,0){\footnotesize (b)}
\end{overpic}
\caption{\small (a) Rank correlation between  integral \cscore and the \cscore for a particular subset ratio, $s$. The peak of each curve indicates the training set size that best reveals generalization of the model.
(b)  Joint distribution of \cscore per-class means and standard deviations on ImageNet. Samples from representative classes (\textcolor{red}{$\star$}'s) are shown in Figure~\ref{fig:imagenet-per-class-egs}.}\label{fig:MS_IPC}
\end{figure}

Figure~\ref{fig:MS_IPC}a shows the rank correlation between the integral score and the score for
a given $s$, as a function of $s$ for our three smaller data sets, MNIST, CIFAR-10, and CIFAR-100. 
Examining the green CIFAR-10 curve, there is a
peak at $s=30$,
indicating that $s=30$ yields the best point-estimate approximation for the integral \cscore.
That the peak is at an intermediate $s$ is consistent with the observation from Figure~\ref{fig:c-profile-3data}
that the \cscore bunches together instances for low and high $s$.
For MNIST (blue curve), a less challenging data set than CIFAR-10, the peak is lower, at $s=10$; for CIFAR-100
(orange curve), a more challenging data set than CIFAR-10, the peak is higher, at
$s=40$ or $s=50$. Thus, the peak appears to shift to larger $s$ for more challenging
data sets. This finding is not surprising: more challenging data sets require a greater
diversity of training instances in order to observe generalization.

Based on these observations, we picked $s=70$ for a point estimate on ImageNet. In particular, 
we train 2,000 ResNet-50 models each with a random 70\% subset of the ImageNet training set, and
estimate the \cscore based on those models.
\begin{figure*}[b]
    \centering
    \includegraphics[width=.17\linewidth]{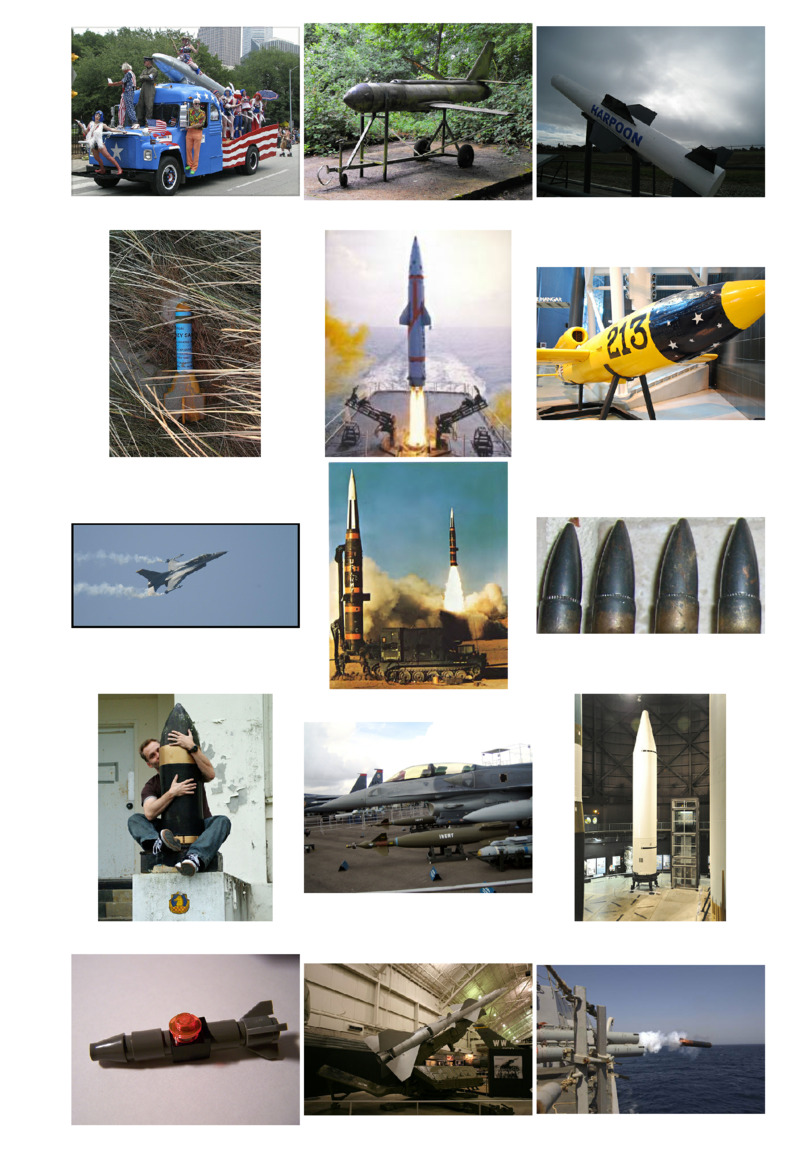}
    \includegraphics[width=.17\linewidth]{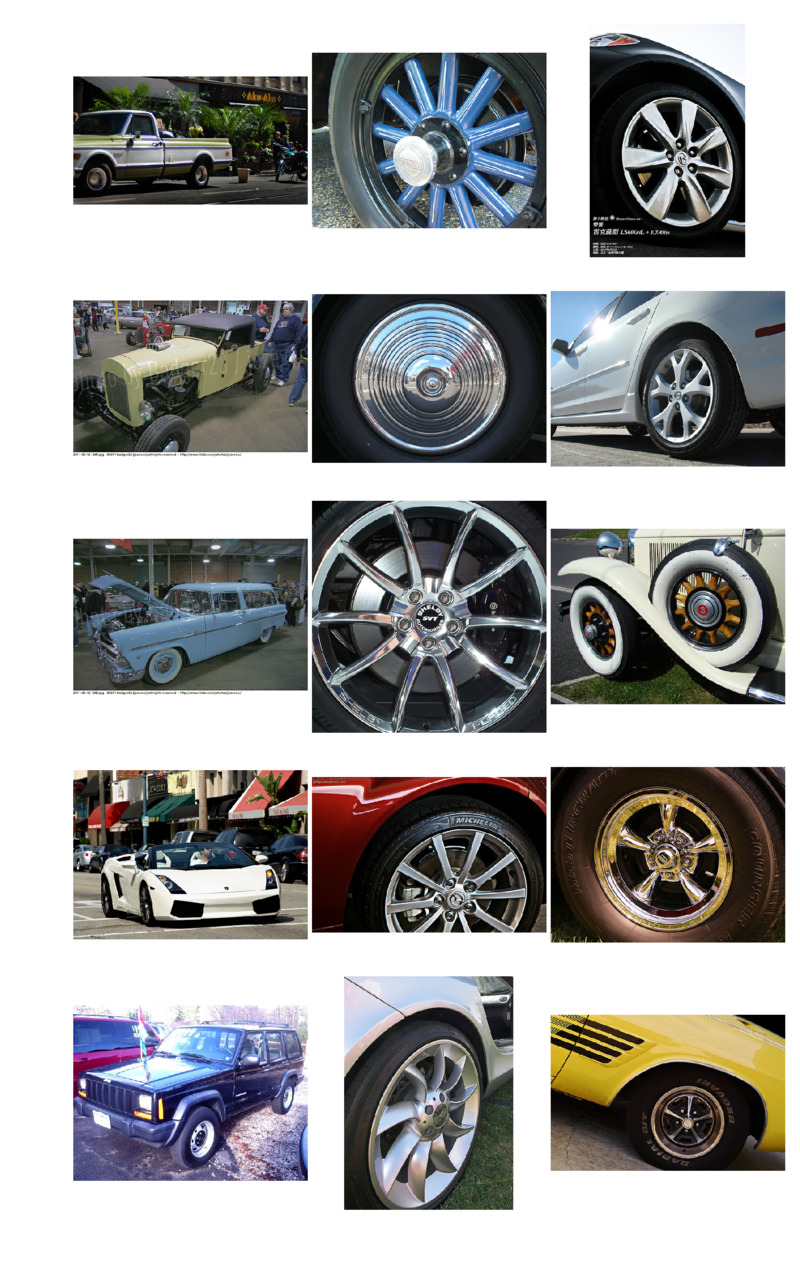}
    \includegraphics[width=.17\linewidth]{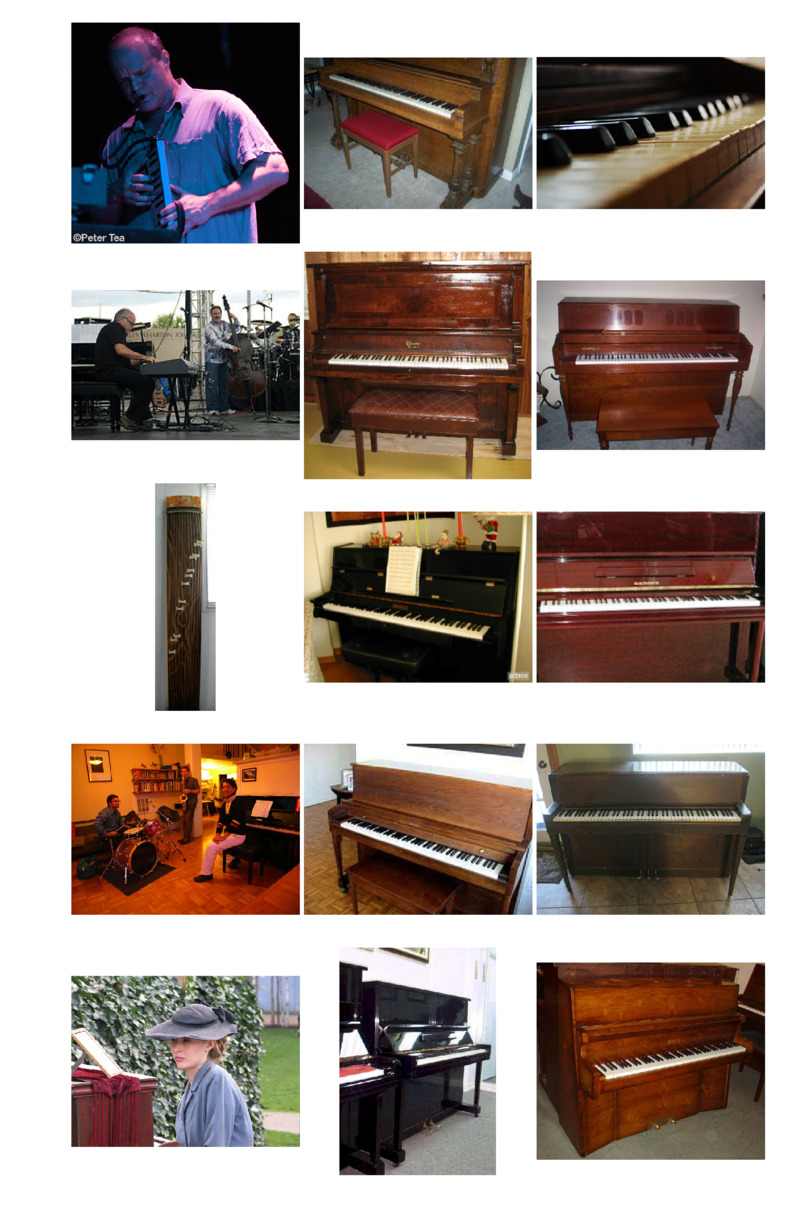}
    \includegraphics[width=.17\linewidth]{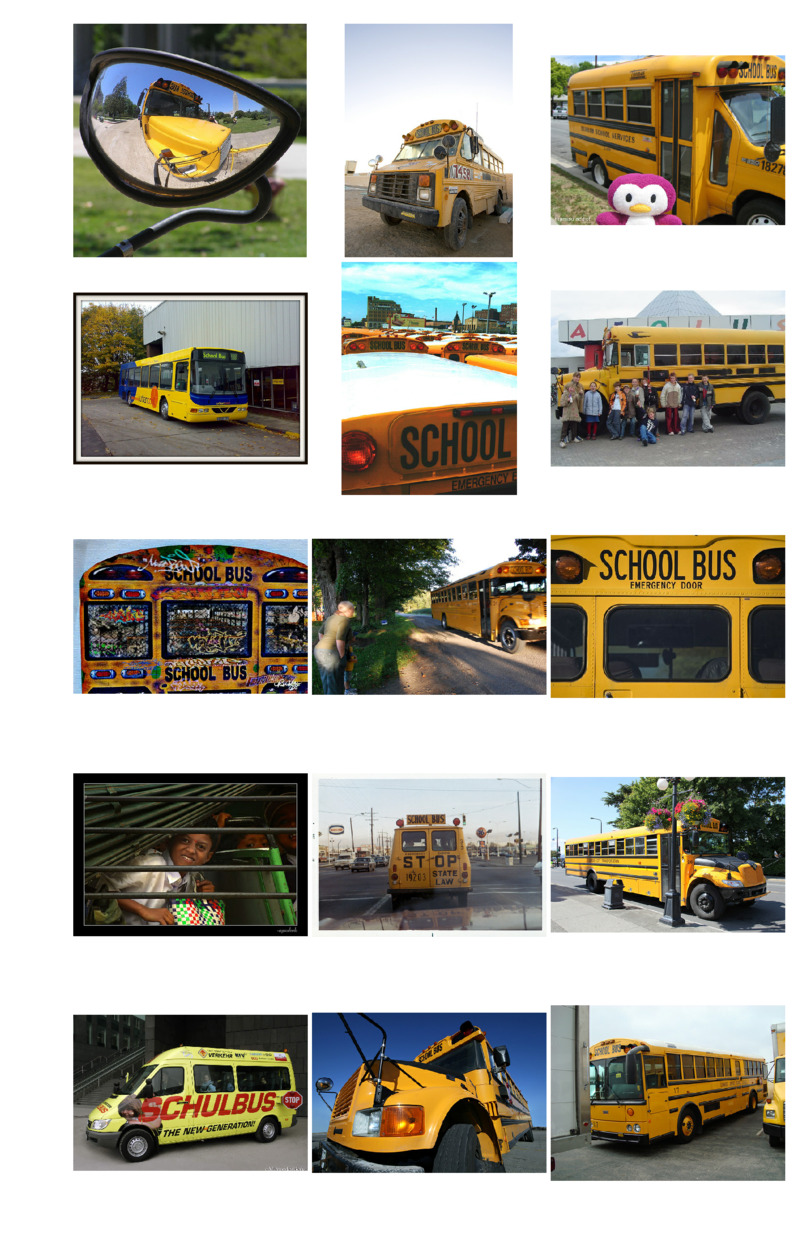}
    \includegraphics[width=.17\linewidth]{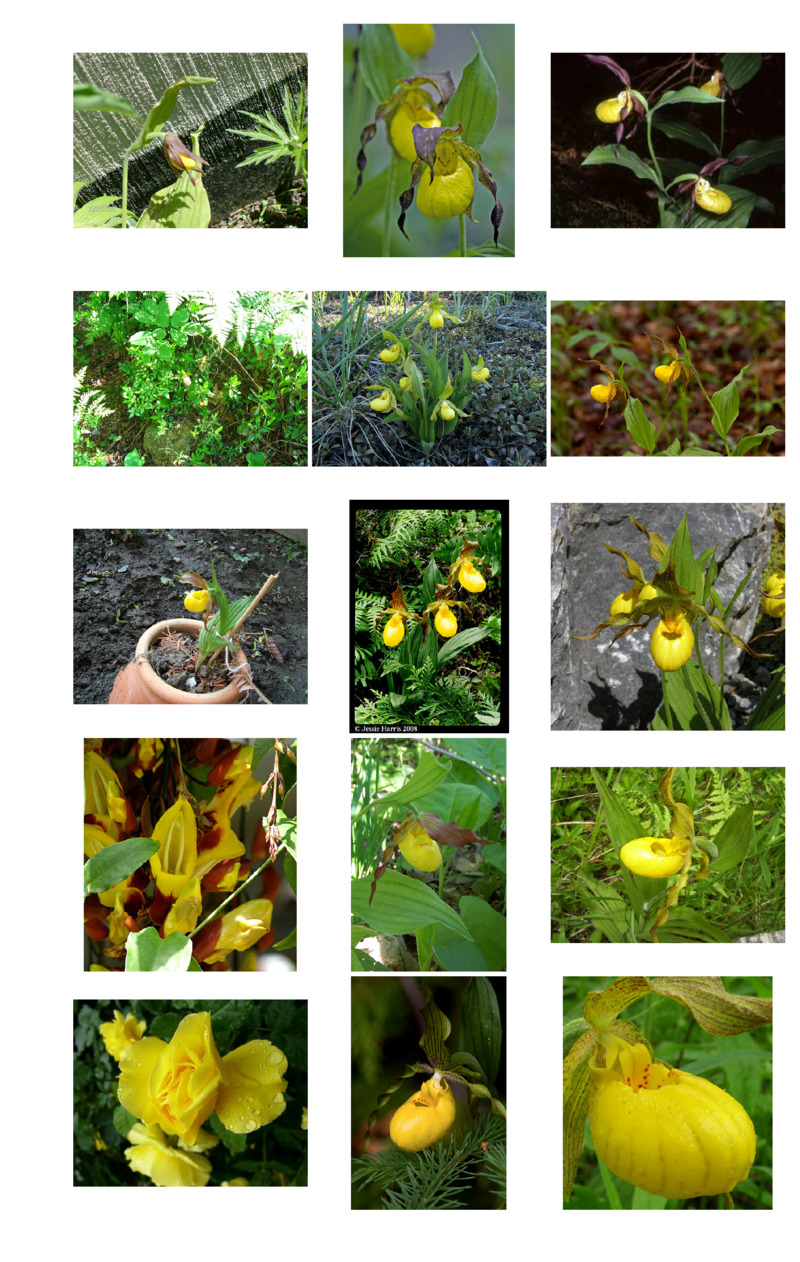}
    \includegraphics[width=.175\linewidth]{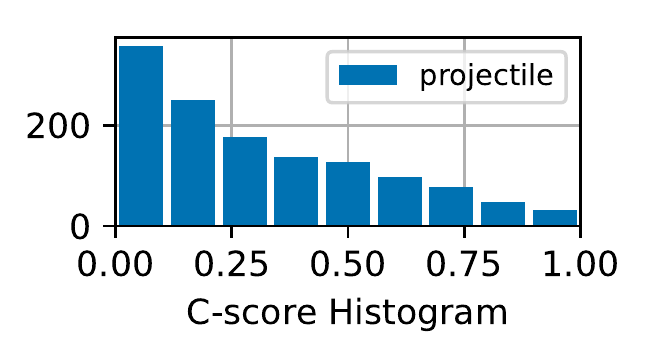}
    \includegraphics[width=.175\linewidth]{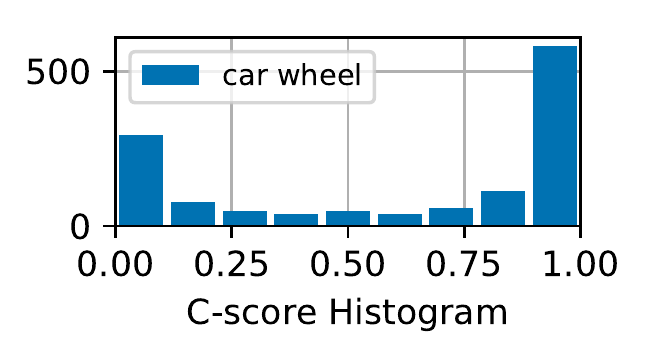}
    \includegraphics[width=.175\linewidth]{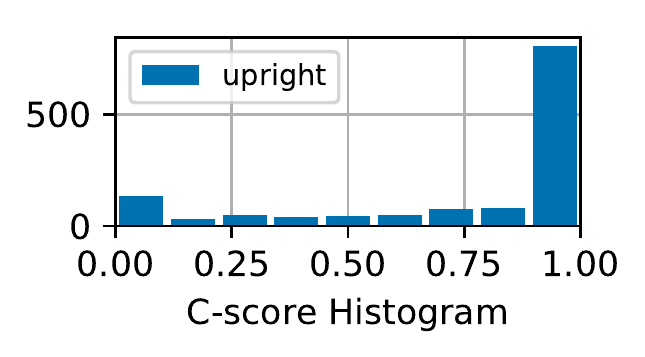}
    \includegraphics[width=.175\linewidth]{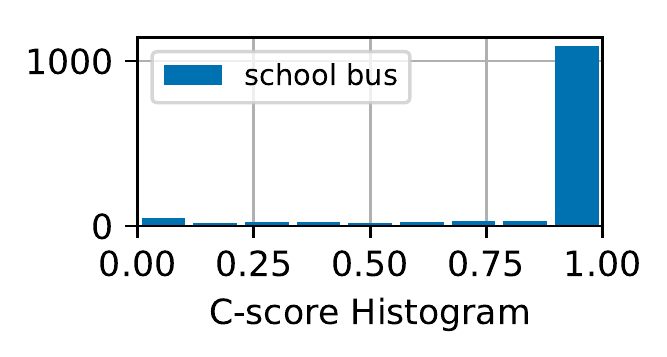}
    \includegraphics[width=.175\linewidth]{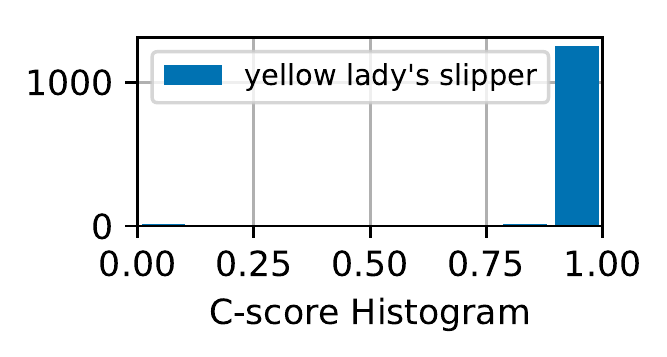}
    \vskip-6pt
    \caption{\small Example images from ImageNet. The 5 classes are chosen to have representative per-class \cscore mean--standard-deviation profiles, as shown in Figure~\ref{fig:MS_IPC}a. For each class, the three columns show sampled images from the (\cscore ranked) top 99\%, 35\%, and 1\% percentiles, respectively. The bottom pane shows the histograms of the \cscores in each of the 5 classes.}
    \label{fig:imagenet-per-class-egs}
\end{figure*}

The examples shown in Figure~\ref{fig:chairs}c are ranked according to this \cscore estimate.
Because ImageNet has 1,000 classes, we cannot offer a simple overview over the entire data set as
in MNIST and CIFAR. Instead, we focus on analyzing the behaviors of individual classes. Specifically,
we compute the mean and standard deviation (SD) of the \cscores of all the examples in a particular
class. The mean \cscores indicates the relative difficulty of classes, and the SD
indicates the diversity of examples within each class. The two-dimensional histogram in Figure~\ref{fig:MS_IPC}a depicts the joint
distribution of mean and SD across all classes.
We selected several classes with various combinations of mean and SD, indicated by
the {\textcolor{red}{$\star$}}'s in Figure~\ref{fig:MS_IPC}a. We then selected
sample images from the top 99\%, 35\% and 1\%  percentile ranked by the \cscore within each class,
and show them in Figure~\ref{fig:imagenet-per-class-egs}.

\emph{Projectile} and \emph{yellow lady's slipper} represent two extreme cases of diverse and unified classes, respectively.
Most other classes lie in the high density region of the 2D histogram in Figure~\ref{fig:MS_IPC}b, and share a common pattern
of a densely populated mode of highly regular examples and a tail of rare, ambiguous examples. The tail becomes smaller from the class \emph{car wheel} to \emph{upright}
and \emph{school bus}.

\section{\cscore Proxies}

We are able to reduce the cost of estimating \cscores from infeasible to feasible, but the procedure is still
very expensive. Ideally, we would like to have more efficient \emph{proxies} that do not require training multiple models.
We use the term \emph{proxy} to refer to any quantity that is well correlated with the \cscore but
does not have a direct mathematical relation to it, as contrasted with \emph{approximations} that
are designed to mathematically approximate the \cscore (e.g., approximating the expectation with empirical averaging).
The possible candidate set for \cscore proxies is very large, as any measure that reflects information about difficulty or regularity of examples could be considered. Our Related Work section mentions a few such possibilities.
In this paper, we primarily study two variants: \emph{pairwise distance based} proxies and \emph{learning speed based} proxies.

\subsection{Pairwise Distance Based Proxies}

Pairwise distance matches our intuition about consistency very well. In fact, our motivating example in Figure~\ref{fig:chairs}a is illustrated in this way.
Intuitively, an example is consistent with the data distribution if it lies near other examples
having the same label. However, if the example lies far from instances in the same class or
lies near instances of different classes, one might not expect it to generalize. Based on this
intuition, we define a relative local-density score:
\begin{equation}
    \hat{C}^{\pm L}(x,y) = \nicefrac{1}{N}\sum\nolimits_{i=1}^N
    2(\mathbf{1}[y=y_i]-\tfrac{1}{2}) K(x_i,x),
    \label{eq:score-kde-cls-weighted}
\end{equation}
where $K(x,x') = \exp(-\|x-x'\|^2/h^2)$ is an RBF kernel with the bandwidth $h$, and
$\mathbf{1}[\cdot]$ is the indicator function. To evaluate the importance of explicit label information, we study two related scores: $\hat{C}^L$ that uses only same-class examples when estimating the local density, and $\hat{C}$ that uses all the neighbor examples by ignoring the labels.
\begin{align}
    \hat{C}^L(x,y) &= \nicefrac{1}{N}\sum\nolimits_{i=1}^N
    \mathbf{1}[y=y_i] K(x_i,x),
    \label{eq:score-kde-cls-masked}\\
    \hat{C}(x) &=
    \nicefrac{1}{N}\sum\nolimits_{i=1}^N K(x_i, x).
    \label{eq:score-kde}
\end{align}
We also study a proxy based on the local outlier factor (LOF) algorithm \citep{Breunig2000}, which measures the local deviation of each point with respect to its neighbours. Since large LOF scores indicate outliers, we use the negative LOF score as a \cscore proxy, denoted by $\hat{C}^\suplof(x)$.

\begin{table}
    \centering
    \caption{Rank correlation between \cscore and pairwise distance based proxies on inputs. Measured with Spearman's $\rho$ and Kendall's $\tau$ rank correlations, respectively.}\vskip3pt
    \label{tab:kde-inputs}\small
    \begin{tabular}{ccccccc}
    \toprule
    & & $\hat{C}$ & $\hat{C}^L$ & $\hat{C}^{\pm L}$ & $\hat{C}^\suplof$ \\
    \midrule
    \multirow{2}{*}{$\rho$}
        & CIFAR-10 & $-0.064$ & $-0.009$ & $0.083$ & $0.103$ \\
    & CIFAR-100 & $-0.098$ & $0.117$ & $0.105$ & $0.151$ \\
    \midrule
            \multirow{2}{*}{$\tau$}
        & CIFAR-10  & $-0.042$ & $-0.006$ & $0.055$ & $0.070$ \\
    & CIFAR-100 & $-0.066$ & $0.078$ & $0.070$ & $0.101$ \\
    \bottomrule
    \end{tabular}
                                                                                                                                                                    \end{table}

\begin{figure}
    \centering
    \begin{overpic}[width=.35\linewidth]{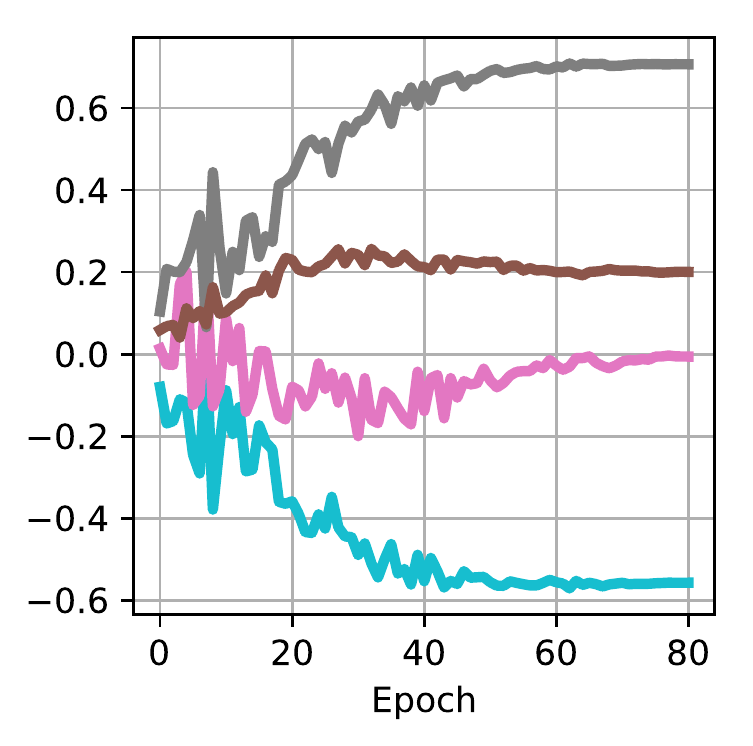}
    \put(-1,3){\scriptsize\textbf{a)} CIFAR-10}
    \end{overpic}
    \begin{overpic}[width=.35\linewidth]{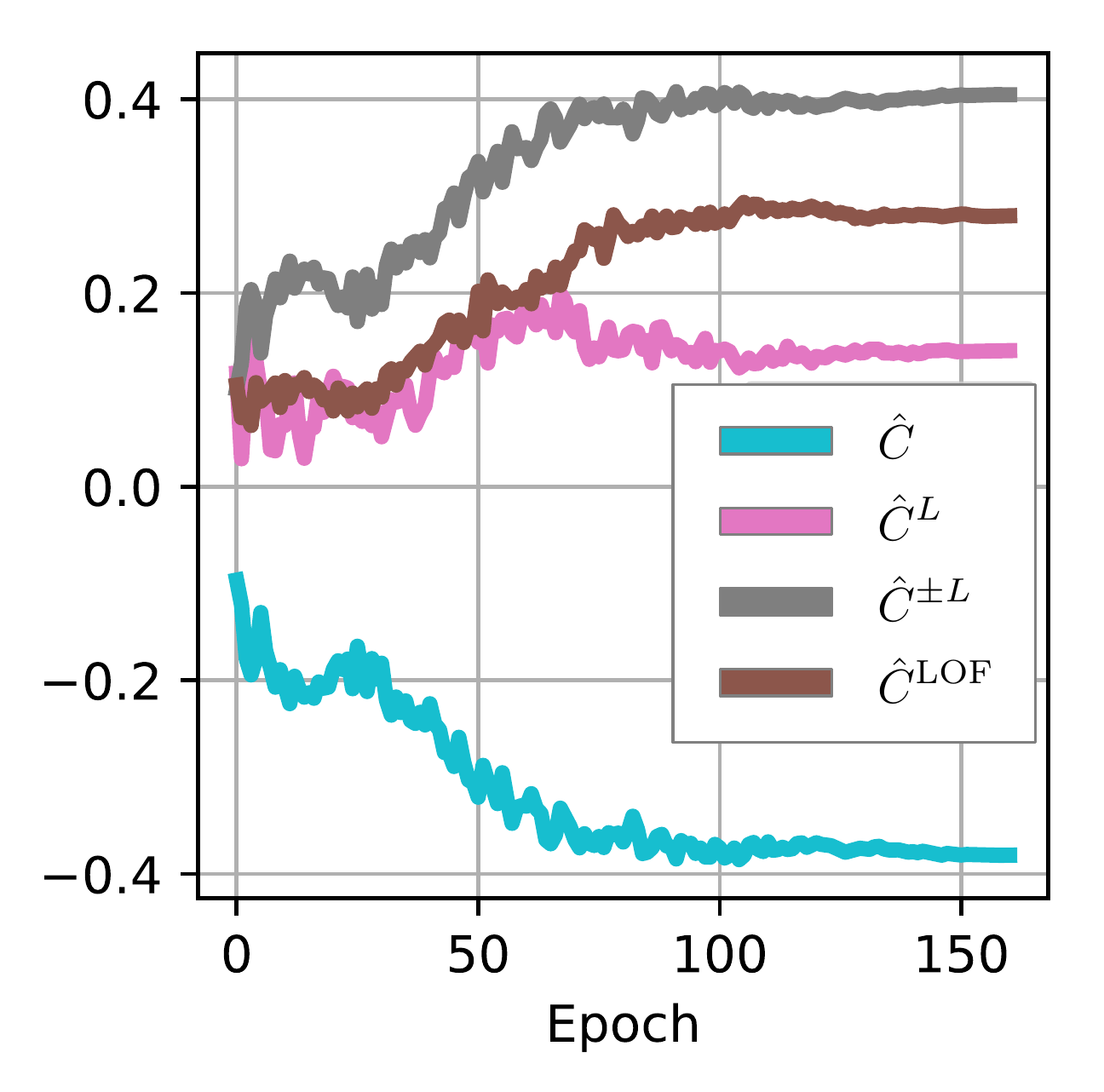}
    \put(-3,3){\scriptsize\textbf{b)} CIFAR-100}
    \end{overpic}\hfill
    \begin{overpic}[width=.260\linewidth]{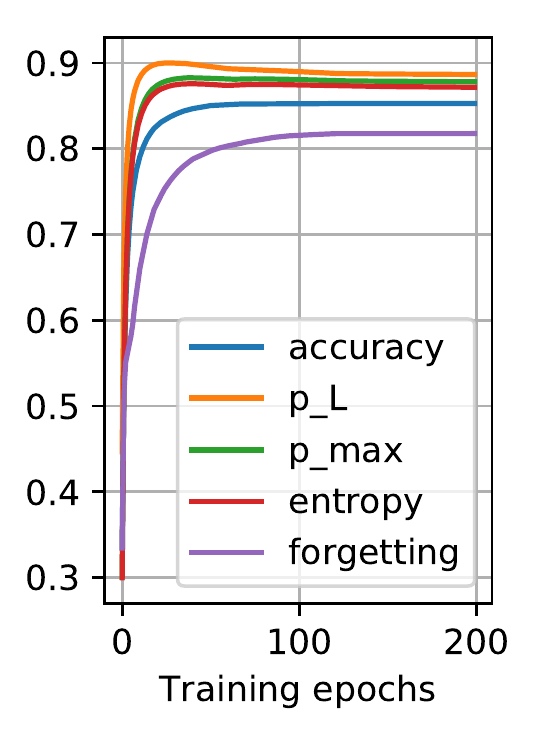}
    \put(1,3){\scriptsize\textbf{c)}}
    \end{overpic}\vspace{-3pt}
    \caption{\small (a-b) Spearman rank correlation between \cscore and distance based proxies using learned hidden representations. (c) Spearman rank correlation between \cscore and learning speed based proxies on CIFAR-10.}
    \label{fig:repr-kde-spearmanr}
\end{figure}

Table~\ref{tab:kde-inputs} shows the agreement between the proxy scores and the estimated \cscore. Agreement
is quantified by two rank correlation measures on three data sets. $\hat{C}^\suplof$ performs slightly better than the other proxies, but none of the proxies has high enough correlation to be useful, because it is very hard to obtain semantically meaningful distance estimations from the raw pixels.

We further evaluate the proxies
using the penultimate layer of the network as a representation of an image:
$\smash{\hat{C}_h^{\pm L}}$, $\smash{\hat{C}_h^L}$, $\smash{\hat{C}_h}$ and $\smash{\hat{C}^\suplof_h}$, with the subscript $h$ indicating distance 
in hidden space. In particular, we train neural network models with the same specification on the full training set. We plot the correlation between the \cscore and the proxy based on the learned representation at each epoch as a function of training epoch in Figure~\ref{fig:repr-kde-spearmanr}a,b.
For both data sets, the proxy score that correlates best with the \cscore is $\smash{\hat{C}_h^{\pm L}}$ (\textcolor[rgb]{0.5,0.5,0.5}{grey}), followed by
$\smash{\hat{C}^\suplof_h}$ (\textcolor[rgb]{0.55,0.34,0.29}{brown}), then $\smash{\hat{C}_h^L}$ (\textcolor[rgb]{0.9,0.47,0.76}{pink}) and $\smash{\hat{C}_h}$ (\textcolor[rgb]{0.1,0.75,0.81}{blue}).
Clearly, appropriate use of labels helps with the ranking. 

The results reveal interesting properties of the hidden representation. One might be concerned that as training progresses, the representations will optimize toward
the classification loss and may discard inter-class relationships that could be potentially useful for other downstream tasks \citep{Scott2018}. However, our
results suggest that $\smash{\hat{C}_h^{\pm L}}$ does not diminish as a predictor of the \cscore, even long after training converges. Thus, at least some information concerning
the relation between different examples is retained in the representation, even though intra- and inter-class similarity is not very relevant for a classification model.
To the extent that the hidden representation---crafted through a discriminative loss---preserves class structure, one might expect that the \cscore could be predicted without
label reweighting; however, the poor performance of  $\smash{\hat{C}_h}$ suggests otherwise.

\begin{figure}
    \hfill
    \includegraphics[width=0.15\linewidth]{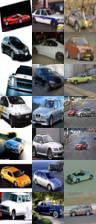}
    \includegraphics[width=0.15\linewidth]{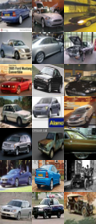}
    \includegraphics[width=.15\linewidth]{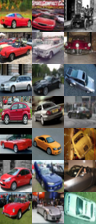}\hspace{1pt}     \hfill 
    \includegraphics[width=0.15\linewidth]{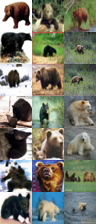}
    \includegraphics[width=0.15\linewidth]{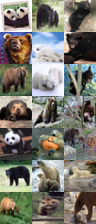}
    \includegraphics[width=.15\linewidth]{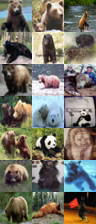}     \vspace{-5pt}
    \caption{\small (Left pane) The 3 blocks show examples from CIFAR-10 ``automobile'' ranked by $\hat{C}^{\pm L}$, $\hat{C}^{\pm L}_h$ and the \cscore, respectively. The three columns in each block shows the top, middle and middle ranked examples, respectively. (Right pane) Examples from CIFAR-100 ``bear'' shown in the same layout.}
    \label{fig:proxy-egs}
\end{figure}

Figure~\ref{fig:proxy-egs} visualizes examples ranked by the class weighted local density scores in the input and learned hidden space, respectively, in comparison with examples ranked by the \cscore. The ranking calculated in the input space relies heavily on low level features that can be derived directly from the pixels like strong silhouette. The rankings calculated from the learned hidden space correlate better with \cscore, though the visualizations show that they could not faithfully detect the dense cluster of highly uniform examples with high \cscores.

In summary, while pairwise distance based proxies are very intuitive to formulate, in practice, the rankings are very sensitive to the underlying distance metrics. 

\subsection{Learning Speed Based Proxies}

Inspired by our observations in the previous section that the speed-of-learning tends to correlate with the \cscore rankings,
we instead focus on a class of learning-speed based proxies that have the added bonus of being trivial to compute.
Intuitively, a training example that is consistent with many others should be learned quickly because the gradient steps for all consistent
examples should be well aligned. One might therefore conjecture that strong regularities in a data set are not only better learned at asymptote---leading
to better generalization performance---but are also learned \emph{sooner} in the time course of training. This \emph{learning speed} hypothesis is
nontrivial, because the \cscore is defined for a held-out instance following training, whereas learning speed is defined for a training instance during training.
This hypothesis is qualitatively verified from Figure~\ref{fig:sgd-vs-adam}. In particular, the \textcolor[rgb]{.25,1,1}{cyan} examples having the lowest \cscores
are learned most slowly and the \textcolor[rgb]{0.85,0.16,1}{purple} examples having the highest \cscores are learned most quickly. Indeed, learning speed is monotonically related to \cscore bin.

Figure~\ref{fig:repr-kde-spearmanr}c shows a quantitative evaluation, where we compute the Spearman's rank correlation between the \cscore of an instance and various proxy scores based on learning speed. In particular, we test \emph{accuracy} (0-1 correctness), $p_L$ (softmax confidence on the correct class), $p_{\max}$ (max softmax confidence across all classes) and \emph{entropy} (negative entropy of softmax confidences). We use \emph{cumulative} statistics which average from the beginning of training to the current epoch because the cumulative statistics yield a more stable measure---and higher correlation---than statistics based on a single epoch.
We also compare to a \emph{forgetting-event} statistic \citep{toneva2018empirical}, which is simply a count of the number of
transitions from ``learned'' to ``forgotten'' during training.
All of our proxies show strong correlation with the \cscore: $p_L$ reaches $\rho\approx 0.9$ at the peak; $p_{\max}$ and \emph{entropy} perform similarly, both slightly worse than $p_L$. 
The \emph{forgetting event} statistic slightly underperforms our proxies and takes a larger number of training epochs to reach its peak correlation. We suspect this is because forgetting events happen only \emph{after} an example is learned, so unlike the proxies studied here, forgetting statistics for hard examples cannot be obtained in the earlier stage of training.

\section{Application}

By characterizing the structural regularities in large scale datasets, the \cscore provides powerful tools for analyzing data sets, learning dynamics, and to diagnose and potentially improve learning systems. In this section, we provide several illustrative applications along this line.

\begin{figure}
    \centering
    \begin{overpic}[width=.67\linewidth]{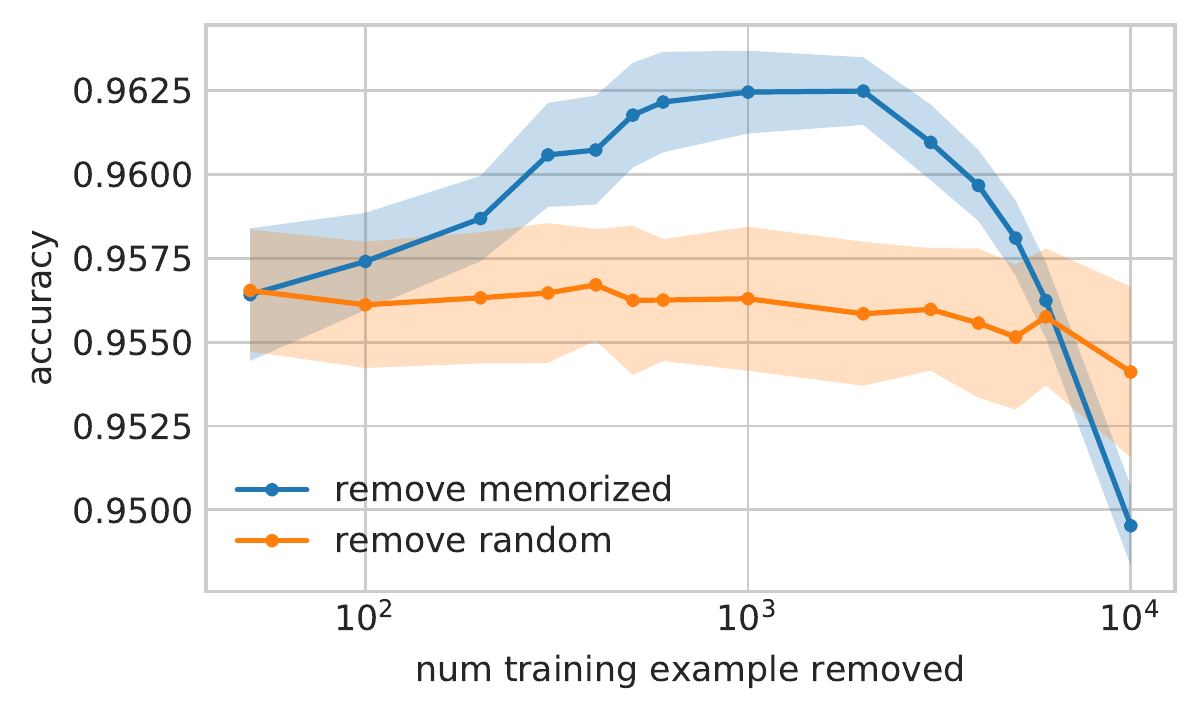}
    \put(0,0){\footnotesize (a)}
    \end{overpic}\hfill
    \begin{overpic}[width=.32\linewidth]{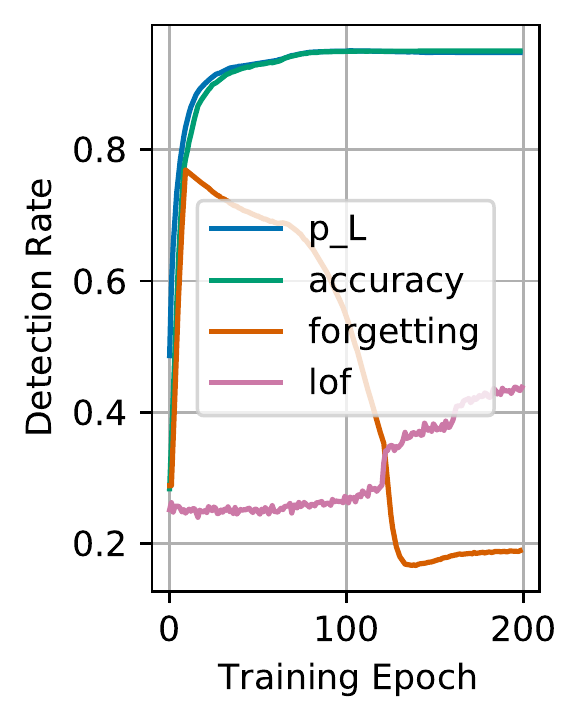}
    \put(0,0){\footnotesize (b)}
    \end{overpic}\vspace{-5pt}
    \caption{(a) Model performance on SVHN when certain number of examples are removed from the training set. (b) Detection rate of label-flipped outliers on CIFAR-10.}
    \label{fig:removal-svhn}
\end{figure}

In the first example, we demonstrate the effects of removing the irregular training examples. In Figure~\ref{fig:removal-svhn}a, we show the performance of models trained on the SVHN~\citep{netzer2011reading} training set as a function of the number of lowest \cscore examples removed. For comparison, we show the performance with the same number of \emph{random} examples removed. We found that the model performance improves as we remove the lowest ranked training examples, but it eventually deteriorates when too many (about $10^4$) training examples are removed. This deterioration occurs because the \cscore typically ranks mislabeled instances toward the bottom, followed by---at least in this data set---correctly labeled but rare instances. Although the mislabeled instances have no utility, the rare instances do, causing a drop in performance as more rare instances are removed. On data sets with fewer mislabelings, such as CIFAR-10,
we did not observe an advantage of removing low-ranked examples versus removing random examples.

To quantitatively evaluate the outlier identification rate, we construct a modified dataset by corrupting a random fraction $\gamma=25\%$ of the CIFAR-10 training set with random label assignments, so that we have the ground-truth indicators for the outliers. We then identify the fraction $\gamma$ with the lowest ranking by our two most promising learning-speed based \cscore proxies---cumulative accuracy and $p_L$. Figure~\ref{fig:removal-svhn}b shows the detection rate---the fraction of the lowest ranked examples which are indeed outliers; the two \cscore proxies successfully identify over 95\% of outliers. This is consistent with previous work~\citep{pleiss2019detecting} showing the loss curves could be informative at detecting noisy examples. We also evaluated the forgetting-event statistic \citep{toneva2018empirical} and the local outlier factor (LOF)~\citep{Breunig2000} algorithm based on distances in the hidden space, but neither is competitive.

\begin{figure}
\begin{overpic}[width=.49\linewidth]{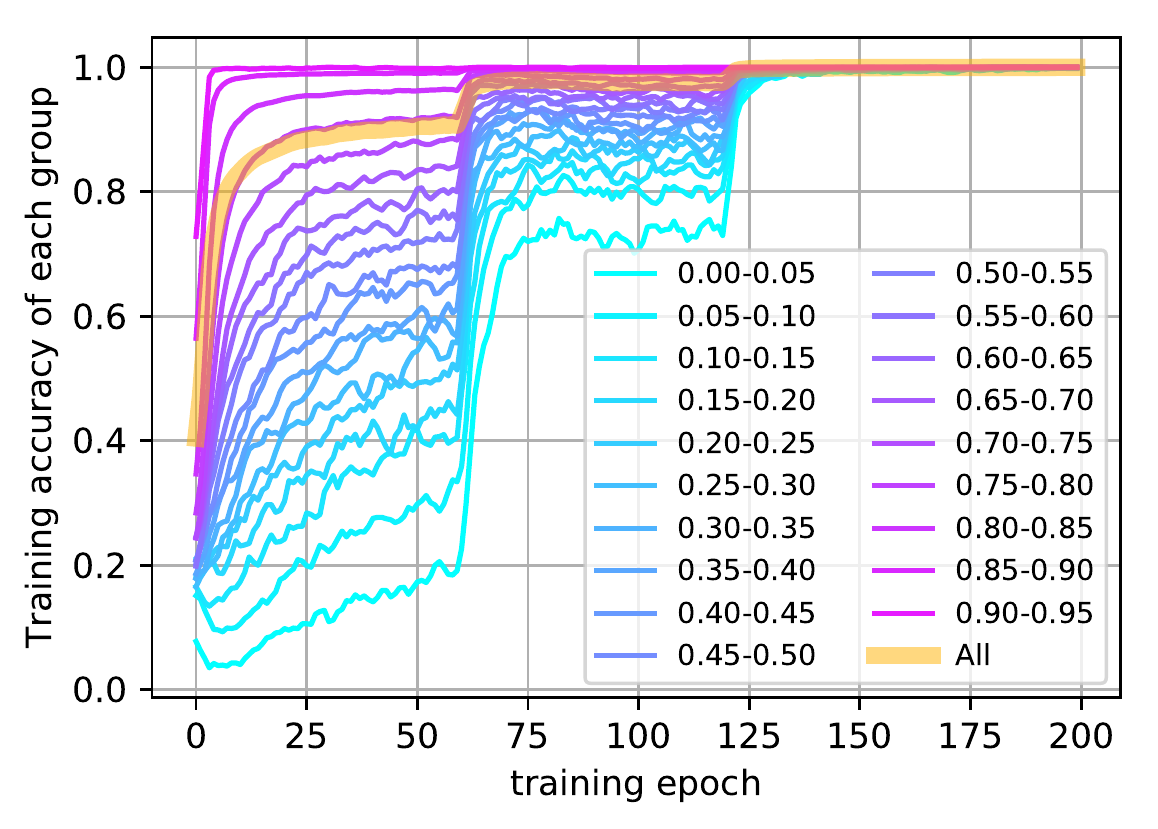}
\put(1,0){\scriptsize\textbf{(a) SGD}}
\end{overpic}
\begin{overpic}[width=.49\linewidth]{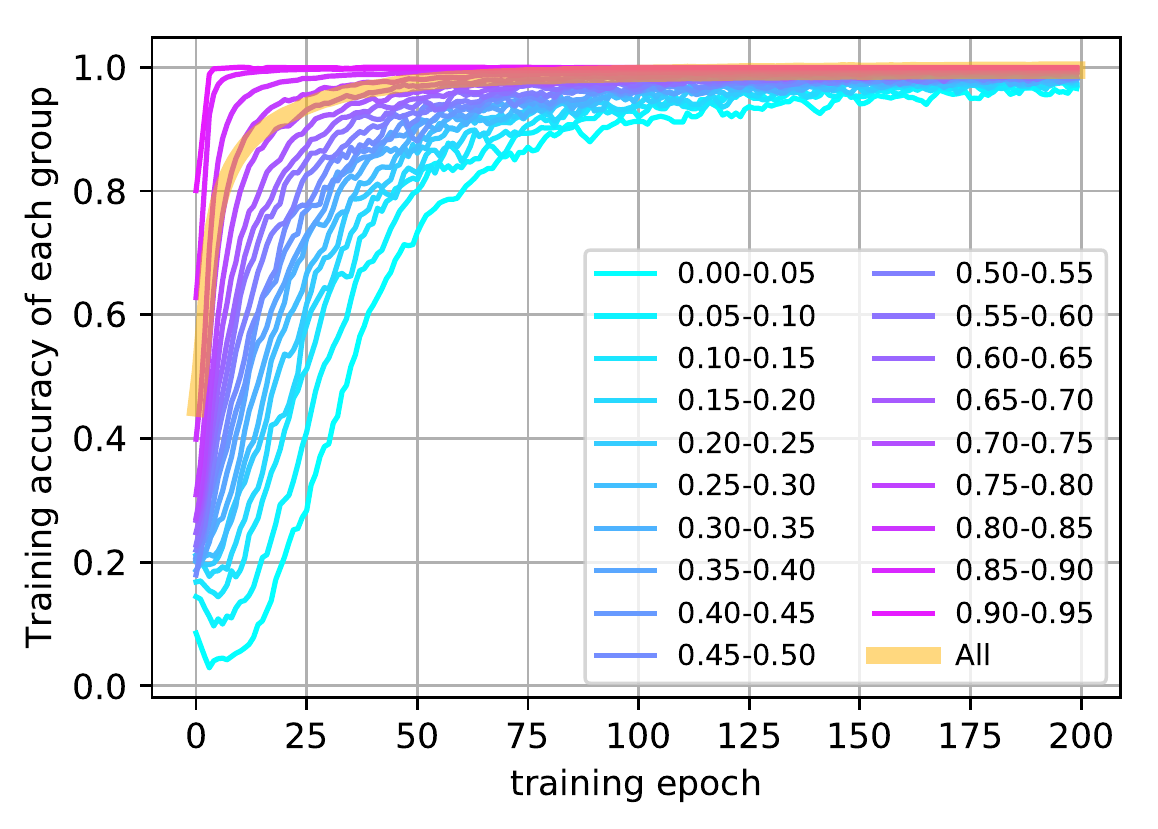}
\put(1,0){\scriptsize\textbf{(b) Adam}}
\end{overpic}
\vskip-10pt
\caption{\small Learning speed of CIFAR-10 examples grouped by \cscore. The thick transparent curve shows the average accuracy over the entire training set. SGD achieves test accuracy 95.14\%, Adam achieves 92.97\%.}\label{fig:sgd-vs-adam}
\end{figure}

In the final example, we demonstrate using the \cscore to study the behavior of different optimizers.
For this study, we partition the CIFAR-10 training set
into subsets by \cscore. Then we record the learning curves---model accuracy over training epochs---for each set.
Figure~\ref{fig:sgd-vs-adam} plots the learning curves for \cscore-binned examples.
The left panel shows SGD training with a stagewise constant learning rate, and the right panel shows the 
Adam optimizer~\citep{adam}, which scales the learning rate adaptively. In both cases, the groups with high 
\cscores (magenta) generally learn faster than the groups with low \cscores (cyan). Intuitively, the high \cscore groups
consist of mutually consistent examples that support one another during training, whereas the low \cscore groups consist
of irregular examples forming sparse modes with fewer consistent peers.  In the case of true outliers, the
model needs to memorize the labels individually  as they do not share structure with any other examples. 

The learning curves have wider dispersion in SGD than in Adam. Early in SGD training where the learning rate is large,
the examples with the lowest \cscores barely learn.
In comparison, Adam shows less spread among the groups and as a result, converges sooner. However, the superior convergence speed of adaptive optimizers
like Adam does not always lead to better generalization \citep{wilson2017marginal,keskar2017improving,luo2019adaptive}. 
We observe this outcome as well: SGD with a stagewise learning rate achieves 95.14\% test accuracy, compared to 92.97\% for Adam. The visualization generated with the help of the \cscore provides an interesting perspective on the difference between the two cases with different generalization performances: SGD with stagewise learning rate effectively enforces a sort of curriculum in which the model focuses on learning the strongest regularities first. This curriculum could help the model building a more solid representation based on domain regularities, when compared to Adam that learns all examples at similar pace.

\section{Discussion}

We formulated a \emph{consistency profile} for individual examples in a data set that reflects the probability of correct generalization to the example as a function of training set size. This profile has strong ties to generalization theory as it essentially measures the per-instance generalization. We distilled the profile
into a scalar \cscore, which provides a total ordering of the instances in a data set by essentially the sample complexity---the amount of training data required---to ensure correct generalization to the instance. By studying the estimated scores on real world datasets, we show that this formulation captures well the basic intuitions about data regularity in both human and machine learning. 

To leverage the \cscore to analyze structural regularities in complex data sets,
we derived a \cscore estimation procedure and obtained \cscores for examples in MNIST, CIFAR-10, CIFAR-100, and ImageNet.
The \cscore estimate helps to characterize the continuum between a densely populated mode consisting of aligned, centrally cropped examples with unified shape and color profiles, and sparsely populated modes of just one or two instances.

We further studied two variants of computationally efficient proxies to the \cscore. We found that the pairwise distance based proxies are sensitive to the underlying distance metrics, while the learning speed based proxies generally provide better correlation with the \cscore. 

We demonstrate examples of potential applications of the \cscore as analytical tools to inspect large scale datasets and the learning systems trained on the data, which provides insights to the otherwise complicated and opaque systems. In particular, we show that the \cscore could be used to identify outliers and provide detailed analysis of the learning dynamics when comparing different optimizers.

One feature of our formulation is that the \cscore depends on the neural network architecture, and more generally on the learning algorithm. Just like how a math major and a music major might have different opinions on the difficulty of courses, different neural networks could have different inductive biases a priori, and the \cscore captures this fact. In practice, we found that the \cscore estimations are consistent among commonly used convolutional networks, potentially because they are not that different from each other. In particular, we compared the Inception based estimation on CIFAR-10 with ResNet-18, VGG-11 and VGG-16, and found the Spearman's $\rho$ correlations are above 0.91. Recently some new \emph{convolution-free} architectures based on attention mechanism~\citep{dosovitskiy2020image} or dense connections~\citep{tolstikhin2021mlp,melas2021you,touvron2021resmlp} emerged and achieved similar performance as their convolutional counterparts on standard image classification benchmarks. We leave it as future work to conduct extensive comparison on more diverse architectures and emerging algorithms such as finetuning after self-supervised learning~\citep{chen2020simple,Chen2020-dk,Grill2020-qw,caron2021emerging}, and so on.

In the 1980s, neural nets were touted for learning \emph{rule-governed behavior} without explicit rules \citep{rumelhart1986}.
At the time, AI researchers were focused on constructing expert systems by extracting explicit rules
from human domain experts.  Expert systems ultimately failed because the diversity and nuance of statistical regularities in a domain was too great
for any human to explicate. In the modern deep learning era, researchers have made much progress in automatically extracting regularities from
data.  Nonetheless, there is still much work to be done to understand these regularities, and how the consistency relationships among instances determine
the outcome of learning. By defining and investigating a consistency score, we hope to have made some progress in this direction. We have released the precomputed \cscores on standard deep learning benchmark datasets to foster future research along this direction.

\section*{Code and Pre-computed \cscores}
\label{app:exported-cscores}

We provide code implementing our C-score estimation algorithms, and
pre-computed \cscores and associated model checkpoints for CIFAR-10, CIFAR-100 and ImageNet
(downloadable from \url{https://pluskid.github.io/structural-regularity/}).
The exported files are in Numpy's data format saved via \texttt{numpy.savez}. For CIFAR-10 and CIFAR-100, the exported file contains two arrays \texttt{labels} and \texttt{scores}. Both arrays are stored in the order of training examples as defined by the original data sets found at \url{https://www.cs.toronto.edu/~kriz/cifar.html}. The data loading tools provided in some deep learning library might not be following the original data example orders, so we provided the \texttt{labels} array for easy sanity check of the data ordering.

For ImageNet, since there is no well defined example ordering, we order the exported scores arbitrarily, and include a script to reconstruct the data set with index information by using the filename of each example to help identify the example-score mapping.

\section*{Acknowledgements}

We thank Vitaly Feldman for guidance on simulation design and framing of the research, Samy Bengio for
general comments and feedback, and Yoram Singer for making the collaboration possible.

\bibliography{cscore}
\bibliographystyle{icml2021}

\appendix
\input{appendix-content}

\end{document}

%% file: appendix-content.tex
\section{Experiment Details}
\label{app:exp-details}

The details on model architectures, data set information and hyper-parameters used in the experiments for empirical estimation of the \cscore can be found in Table~\ref{tab:cscore-exp-details}. We implement our experiment in Tensorflow \citep{tensorflow2015-whitepaper}. The holdout subroutine used in the empirical \cscore estimation is based on the estimator proposed in \citet{fz2019}, and listed in Algorithm~\ref{alg:hold-out}. Most of the training jobs for \cscore estimation are run on single NVidia\textsuperscript{\textregistered} Tesla P100 GPUs. The ImageNet training jobs are run with 8 P100 GPUs using single-node multi-GPU data parallelization.

The experiments on learning speed are conducted with ResNet-18 on CIFAR-10, trained for 200 epochs while batch size is 32. For optimizer, we use the SGD with the initial learning rate 0.1, momentum 0.9 (with Nesterov momentum) and weight decay is 5e-4. The stage-wise constant learning rate scheduler decrease the learning rate at the 60th, 90th, and 120th epoch with a decay factor of 0.2.

\begin{algorithm}
\caption{Estimation of $\hat{C}_{\hat{\mathcal{D}},n}$}
\label{alg:hold-out}
\begin{algorithmic}
\REQUIRE Data set $\hat{\mathcal{D}}=(X,Y)$ with $N$ examples
\REQUIRE $n$: number of instances used for training
\REQUIRE $k$: number of subset samples
\ENSURE $\hat{C}\in\RR^N$: $(\hat{C}_{\hat{\mathcal{D}},n}(x,y))_{(x,y) \in\hat{\mathcal{D}}}$
\STATE Initialize binary mask matrix $M\leftarrow 0^{k\times N}$
\STATE Initialize 0-1 loss matrix $L\leftarrow 0^{k\times N}$
\FOR{$i\in (1,2,\ldots,k)$}
  \STATE Sample $n$ random indices $I$ from $\{1,\ldots,N\}$
  \STATE $M[i, I]\leftarrow 1$
  \STATE Train $\hat{f}$ from scratch with the subset $X[I],Y[I]$
  \STATE $L[i,:]\leftarrow \mathbf{1}[\hat{f}(X) \neq Y]$
\ENDFOR
\STATE Initialize score estimation vector $\hat{C}\leftarrow 0^{N}$
\FOR{$j\in (1,2,\ldots,N)$}
    \STATE $Q \leftarrow \lnot M[:, j]$
    \STATE $\hat{C}[j] \leftarrow \text{sum}(\lnot L[:, Q]) / \text{sum}(Q)$
\ENDFOR
\end{algorithmic}
\end{algorithm}

\begin{table*}
\caption{Details for the experiments used in the empirical estimation of the \cscore.}
\label{tab:cscore-exp-details}
\centering
  \begin{tabular}{lcccc}
    \toprule
      & MNIST & CIFAR-10 & CIFAR-100 & ImageNet \\
    \midrule
    Architecture & MLP(512,256,10) & Inception$^\dagger$ & Inception$^\dagger$ & ResNet-50 (V2) \\
    Optimizer & SGD & SGD & SGD & SGD \\
    Momentum & 0.9 & 0.9 & 0.9 & 0.9 \\
    Base Learning Rate & 0.1 & 0.4 & 0.4 & 0.1$\times$7 \\
    Learning Rate Scheduler & $\wedge(15\%)^{\star}$ & $\wedge(15\%)^{\star}$ & $\wedge(15\%)^{\star}$ & LinearRampupPiecewiseConstant$^{\star\star}$ \\
    Batch Size & 256 & 512 & 512 & 128$\times$7 \\
    Epochs & 20 & 80 & 160 & 100 \\
    Data Augmentation & \multicolumn{4}{c}{$\cdots\cdots$ Random Padded Cropping$^\circledast$ + Random Left-Right Flipping $\cdots\cdots$} \\
    Image Size & 28$\times$28 & 32$\times$32 & 32$\times$32 & 224$\times$224 \\
    Training Set Size & 60,000 & 50,000 & 50,000 & 1,281,167 \\
    Number of Classes & 10 & 10 & 100 & 1000 \\
    \bottomrule
  \end{tabular}\vskip3pt
  \footnotesize
  \begin{tabular}{rp{.92\linewidth}}
  $\dagger$ & A simplified Inception model suitable for small image sizes, defined as follows: \\
    & \parbox{6em}{\hfill Inception} :: Conv(3$\times$3, 96) $\rightarrow$ Stage1 $\rightarrow$ Stage2 $\rightarrow$ Stage3 $\rightarrow$ GlobalMaxPool $\rightarrow$ Linear.\\
    & \parbox{6em}{\hfill Stage1} :: Block(32, 32) $\rightarrow$ Block(32, 48) $\rightarrow$ Conv(3$\times$3, 160, Stride=2). \\
    & \parbox{6em}{\hfill Stage2} :: Block(112, 48) $\rightarrow$ Block(96, 64) $\rightarrow$ Block(80, 80) $\rightarrow$ Block (48, 96) $\rightarrow$ Conv(3$\times$3, 240, Stride=2).\\
    & \parbox{6em}{\hfill Stage3} :: Block(176, 160) $\rightarrow$ Block(176, 160).\\
    & \parbox{6em}{\hfill Block($C_1$, $C_2$)} :: Concat(Conv(1$\times$1, $C_1$), Conv(3$\times$3,$C_2$)). \\
    & \parbox{6em}{\hfill Conv} :: Convolution $\rightarrow$ BatchNormalization $\rightarrow$ ReLU.\\
  $\star$ & $\wedge(15\%)$ learning rate scheduler linearly increase the learning rate from 0 to the \emph{base learning rate} in the first $15\%$ training steps, and then from there linear decrease to 0 in the remaining training steps. \\
  $\star\star$ & LinearRampupPiecewiseConstant learning rate scheduler linearly increase the learning rate from 0 to the \emph{base learning rate} in the first $15\%$ training steps. Then the learning rate remains piecewise constant with a $10\times$ decay at $30\%$, $60\%$ and $90\%$ of the training steps, respectively.\\
  $\circledast$ & Random Padded Cropping pad 4 pixels of zeros to all the four sides of MNIST, CIFAR-10, CIFAR-100 images and (randomly) crop back to the original image size. For ImageNet, a padding of 32 pixels is used for all four sides of the images.
  \end{tabular}
\end{table*}

\section{Time and Space Complexity}

The time complexity of the holdout procedure for empirical estimation of the \cscore is $\mathcal{O}(S(kT+E))$. Here $S$ is the number of subset ratios, $k$ is number of holdout for each subset ratio, and $T$ is the average training time for a neural network. $E$ is the time for computing the score given the $k$-fold holdout training results, which involves elementwise computation on a matrix of size $k\times N$, and is negligible comparing to the time for training neural networks. The space complexity is the space for training a single neural network times the number of parallel training jobs. The space complexity for computing the scores is $\mathcal{O}(kN)$.

For kernel density estimation based scores, the most expensive part is forming the pairwise distance matrix (and the kernel matrix), which requires $\mathcal{O}(N^2)$ space and $\mathcal{O}(N^2d)$ time, where $d$ is the dimension of the input or hidden representation spaces.

\section{More Visualizations of Images Ranked by \cscore}
\label{app:cscore-more-egs}

\begin{figure*}
    \includegraphics[width=.09\linewidth]{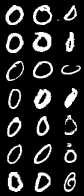}\hfill
    \includegraphics[width=.09\linewidth]{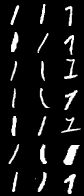}\hfill
    \includegraphics[width=.09\linewidth]{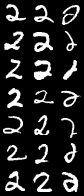}\hfill
    \includegraphics[width=.09\linewidth]{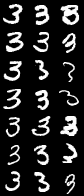}\hfill
    \includegraphics[width=.09\linewidth]{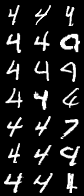}\hfill
    \includegraphics[width=.09\linewidth]{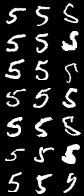}\hfill
    \includegraphics[width=.09\linewidth]{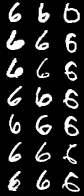}\hfill
    \includegraphics[width=.09\linewidth]{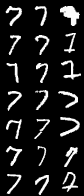}\hfill
    \includegraphics[width=.09\linewidth]{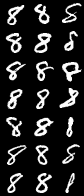}\hfill
    \includegraphics[width=.09\linewidth]{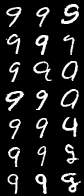}
    \caption{Examples from MNIST. Each block shows a single class; the left, middle, and right columns of a block depict instances with high, intermediate, and low \cscores, respectively.}
    \label{fig:mnist-sweep-full}
\end{figure*}

\begin{figure*}
    \includegraphics[width=.09\linewidth]{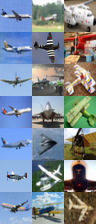}\hfill
    \includegraphics[width=.09\linewidth]{figs/subset-train/cifar10_02-2krank-auc-bigpic-class1}\hfill
    \includegraphics[width=.09\linewidth]{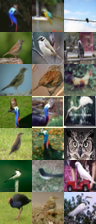}\hfill
    \includegraphics[width=.09\linewidth]{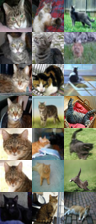}\hfill
    \includegraphics[width=.09\linewidth]{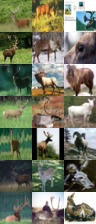}\hfill
    \includegraphics[width=.09\linewidth]{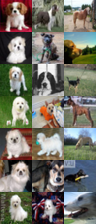}\hfill
    \includegraphics[width=.09\linewidth]{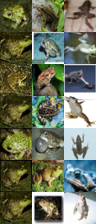}\hfill
    \includegraphics[width=.09\linewidth]{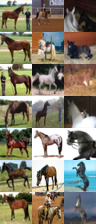}\hfill
    \includegraphics[width=.09\linewidth]{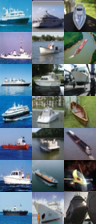}\hfill
    \includegraphics[width=.09\linewidth]{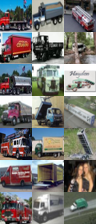}
    \caption{Examples from CIFAR-10. Each block shows a single class; the left, middle, and right columns of a block depict instances with high, intermediate, and low \cscores, respectively.}
    \label{fig:cifar10-sweep-full}
\end{figure*}

\begin{figure*}
    \includegraphics[width=.09\linewidth]{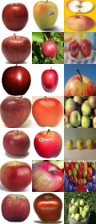}\hfill
    \includegraphics[width=.09\linewidth]{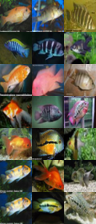}\hfill
    \includegraphics[width=.09\linewidth]{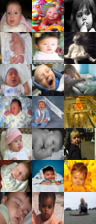}\hfill
    \includegraphics[width=.09\linewidth]{figs/subset-train-cifar100/cifar100_02-2krank-auc-bigpic-class3}\hfill
    \includegraphics[width=.09\linewidth]{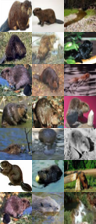}\hfill
    \includegraphics[width=.09\linewidth]{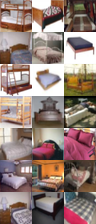}\hfill
    \includegraphics[width=.09\linewidth]{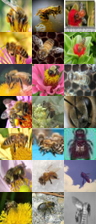}\hfill
    \includegraphics[width=.09\linewidth]{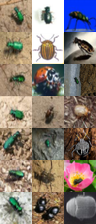}\hfill
    \includegraphics[width=.09\linewidth]{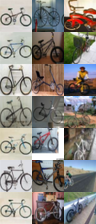}\hfill
    \includegraphics[width=.09\linewidth]{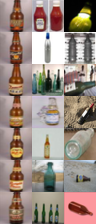}\vskip3px
        \includegraphics[width=.09\linewidth]{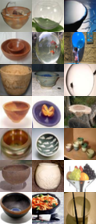}\hfill
    \includegraphics[width=.09\linewidth]{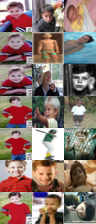}\hfill
    \includegraphics[width=.09\linewidth]{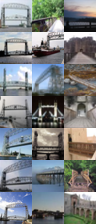}\hfill
    \includegraphics[width=.09\linewidth]{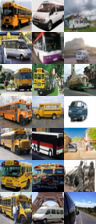}\hfill
    \includegraphics[width=.09\linewidth]{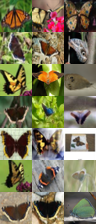}\hfill
    \includegraphics[width=.09\linewidth]{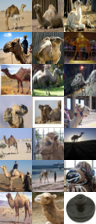}\hfill
    \includegraphics[width=.09\linewidth]{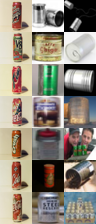}\hfill
    \includegraphics[width=.09\linewidth]{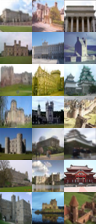}\hfill
    \includegraphics[width=.09\linewidth]{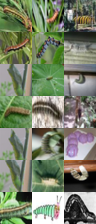}\hfill
    \includegraphics[width=.09\linewidth]{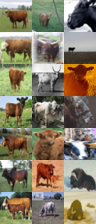}\vskip3pt
        \includegraphics[width=.09\linewidth]{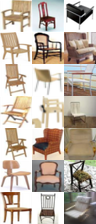}\hfill
    \includegraphics[width=.09\linewidth]{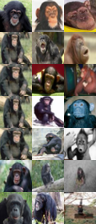}\hfill
    \includegraphics[width=.09\linewidth]{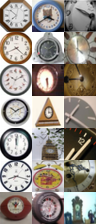}\hfill
    \includegraphics[width=.09\linewidth]{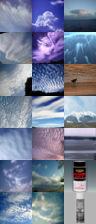}\hfill
    \includegraphics[width=.09\linewidth]{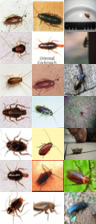}\hfill
    \includegraphics[width=.09\linewidth]{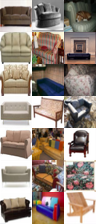}\hfill
    \includegraphics[width=.09\linewidth]{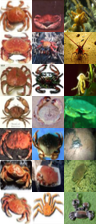}\hfill
    \includegraphics[width=.09\linewidth]{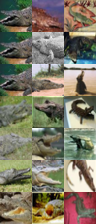}\hfill
    \includegraphics[width=.09\linewidth]{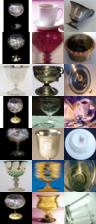}\hfill
    \includegraphics[width=.09\linewidth]{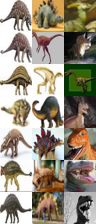}\vskip3pt
        \includegraphics[width=.09\linewidth]{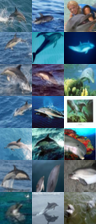}\hfill
    \includegraphics[width=.09\linewidth]{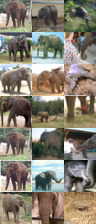}\hfill
    \includegraphics[width=.09\linewidth]{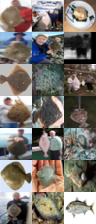}\hfill
    \includegraphics[width=.09\linewidth]{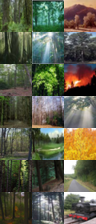}\hfill
    \includegraphics[width=.09\linewidth]{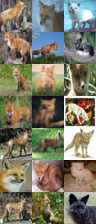}\hfill
    \includegraphics[width=.09\linewidth]{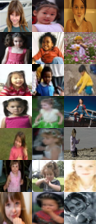}\hfill
    \includegraphics[width=.09\linewidth]{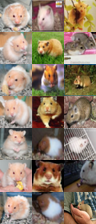}\hfill
    \includegraphics[width=.09\linewidth]{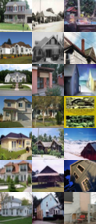}\hfill
    \includegraphics[width=.09\linewidth]{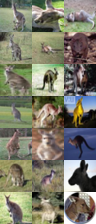}\hfill
    \includegraphics[width=.09\linewidth]{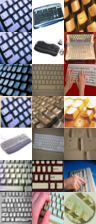}\vskip3pt
        \includegraphics[width=.09\linewidth]{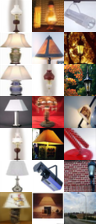}\hfill
    \includegraphics[width=.09\linewidth]{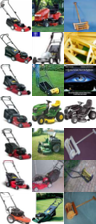}\hfill
    \includegraphics[width=.09\linewidth]{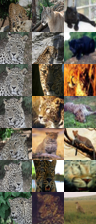}\hfill
    \includegraphics[width=.09\linewidth]{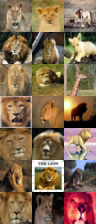}\hfill
    \includegraphics[width=.09\linewidth]{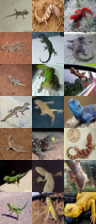}\hfill
    \includegraphics[width=.09\linewidth]{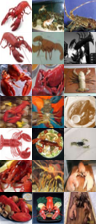}\hfill
    \includegraphics[width=.09\linewidth]{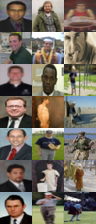}\hfill
    \includegraphics[width=.09\linewidth]{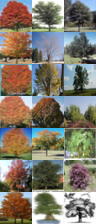}\hfill
    \includegraphics[width=.09\linewidth]{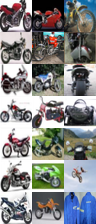}\hfill
    \includegraphics[width=.09\linewidth]{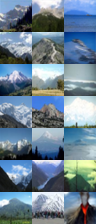}\vskip3pt
        \includegraphics[width=.09\linewidth]{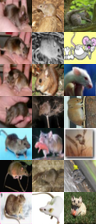}\hfill
    \includegraphics[width=.09\linewidth]{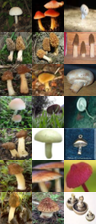}\hfill
    \includegraphics[width=.09\linewidth]{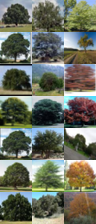}\hfill
    \includegraphics[width=.09\linewidth]{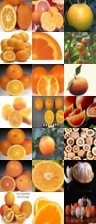}\hfill
    \includegraphics[width=.09\linewidth]{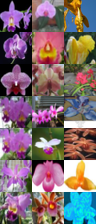}\hfill
    \includegraphics[width=.09\linewidth]{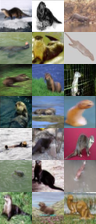}\hfill
    \includegraphics[width=.09\linewidth]{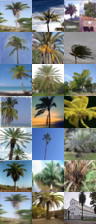}\hfill
    \includegraphics[width=.09\linewidth]{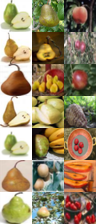}\hfill
    \includegraphics[width=.09\linewidth]{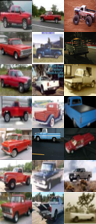}\hfill
    \includegraphics[width=.09\linewidth]{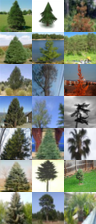}\vskip-2pt
    \caption{Examples from CIFAR-100. Each block shows a single class; the left, middle, and right columns of a block depict instances with high, intermediate, and low \cscores, respectively. The first 60 (out of the 100) classes are shown.}
    \label{fig:cifar100-sweep-full}
\end{figure*}

\begin{figure*}
    \centering
    \includegraphics[width=.18\linewidth]{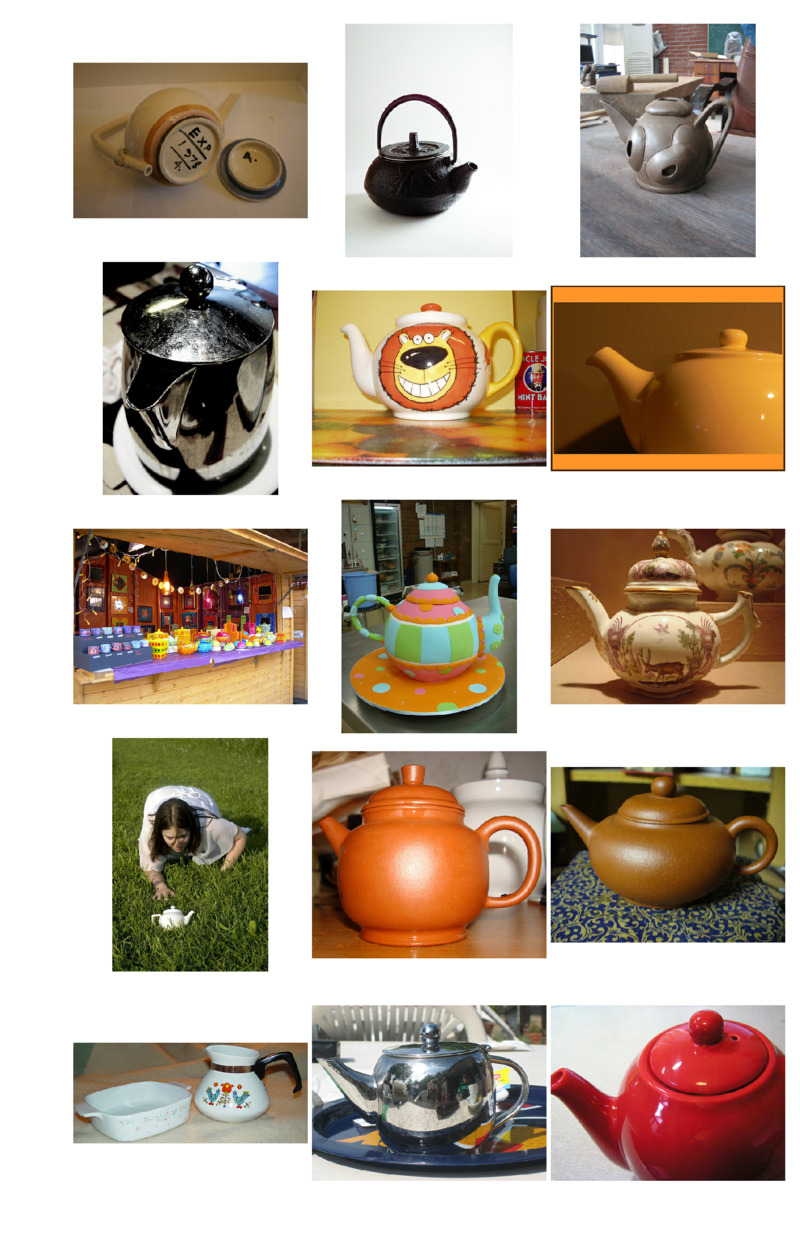}
    \includegraphics[width=.18\linewidth]{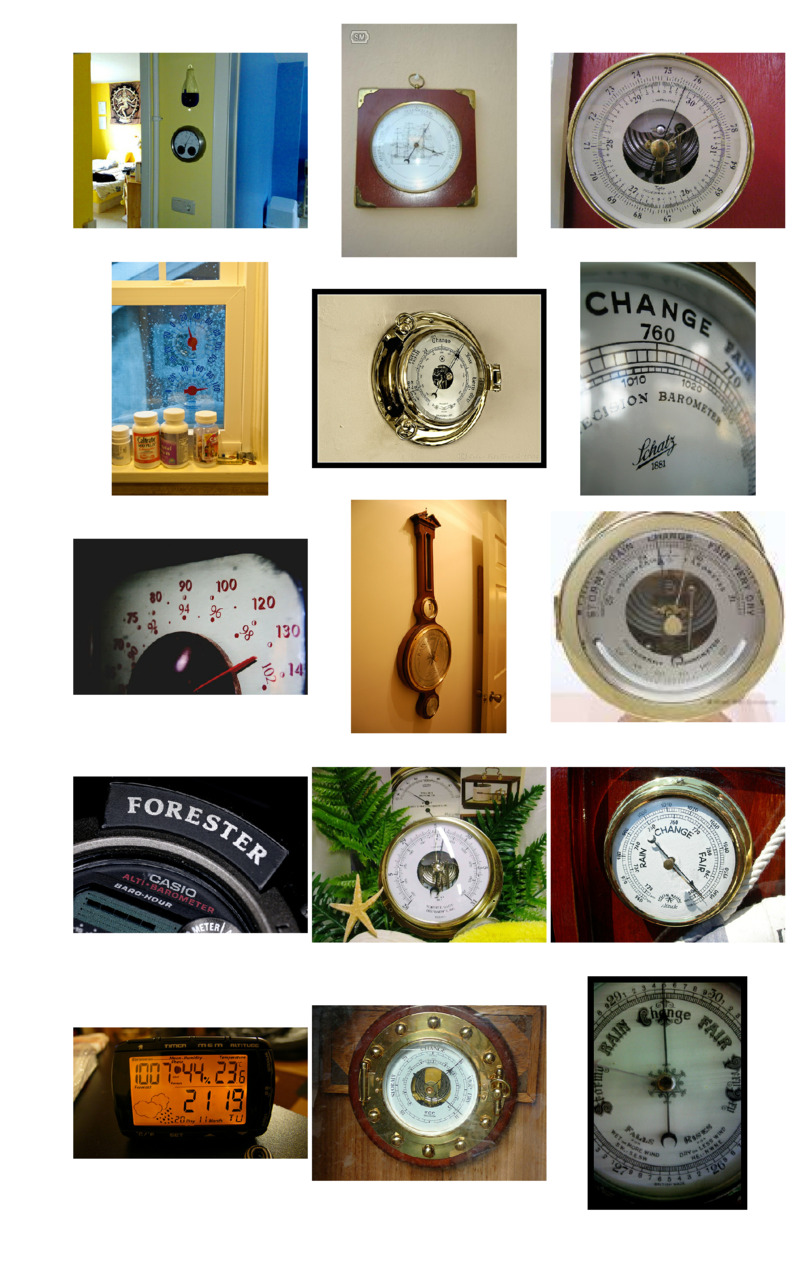}
    \includegraphics[width=.18\linewidth]{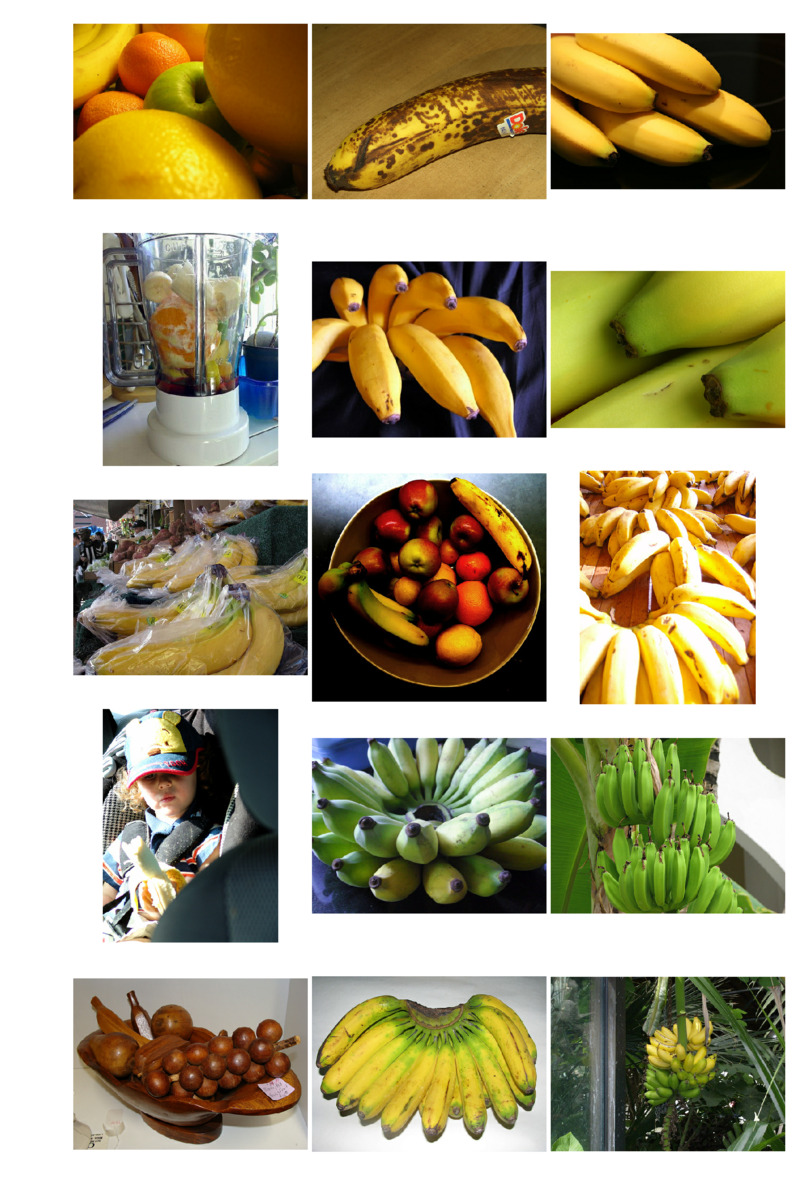}
    \includegraphics[width=.18\linewidth]{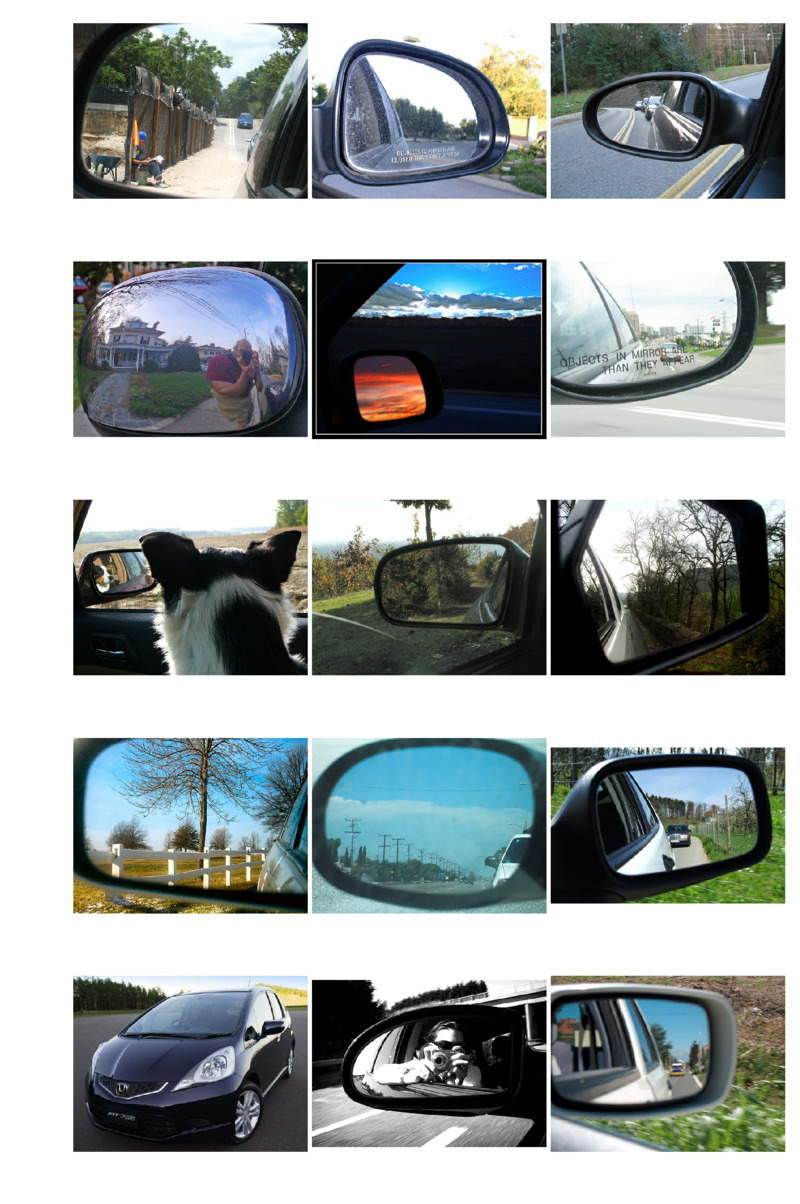}
    \includegraphics[width=.18\linewidth]{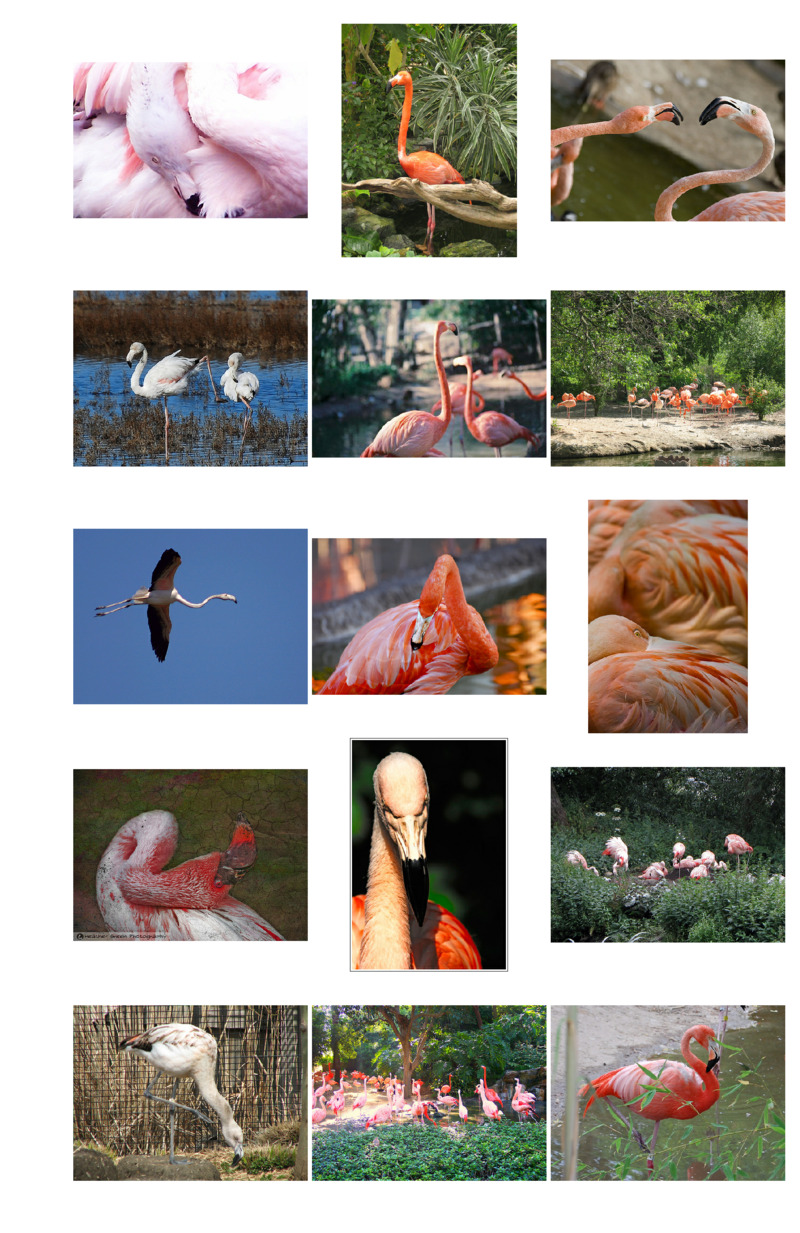}
    \includegraphics[width=.185\linewidth]{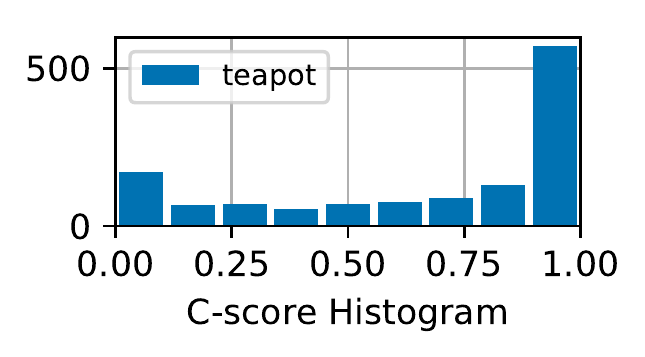}
    \includegraphics[width=.185\linewidth]{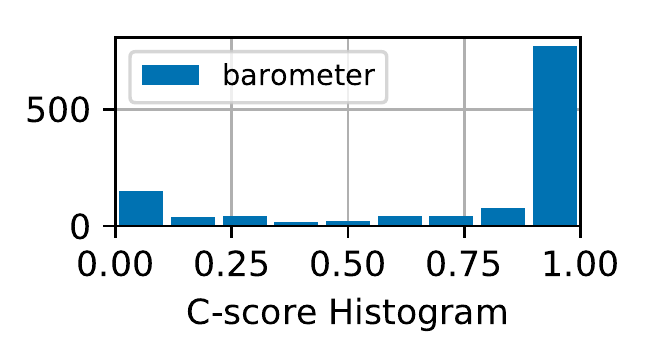}
    \includegraphics[width=.185\linewidth]{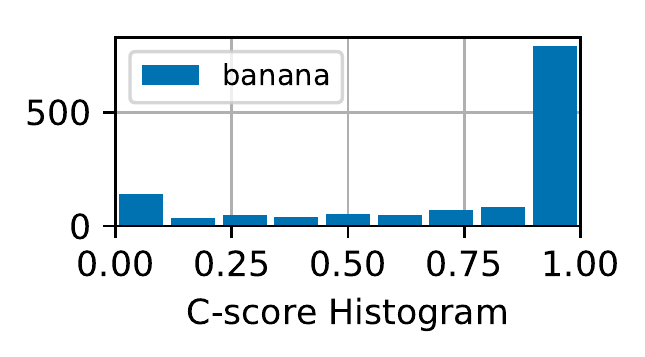}
    \includegraphics[width=.185\linewidth]{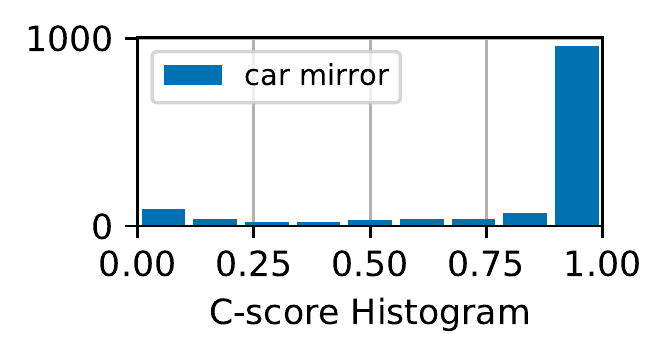}
    \includegraphics[width=.185\linewidth]{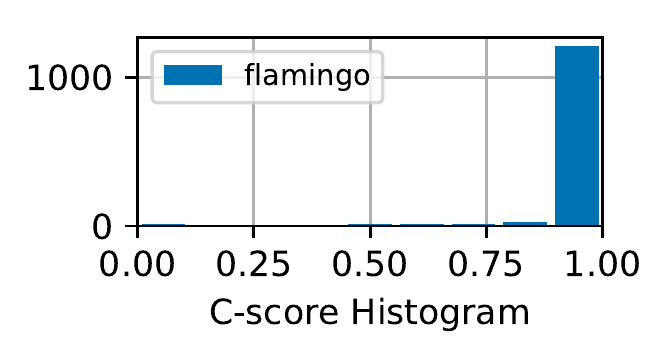}
    \vskip-6pt
    \caption{\small Example images from ImageNet. For each class, the three columns show sampled images from the (\cscore ranked) top 99\%, 35\%, and 1\% percentiles, respectively. The bottom pane shows the histograms of the \cscores in each of the 5 classes.}
    \label{fig:imagenet-per-class-egs-more}
\end{figure*}

\begin{figure*}
    \centering
    \includegraphics[width=.18\linewidth]{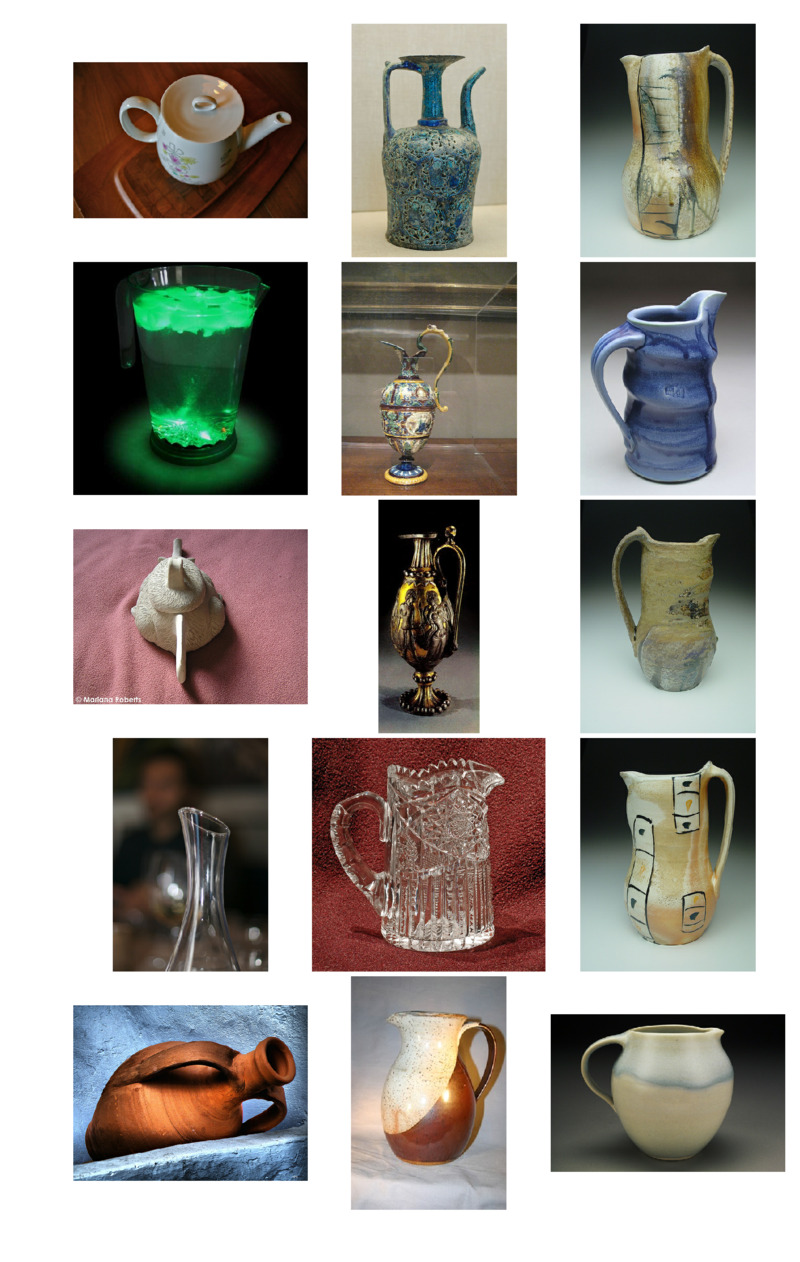}
    \includegraphics[width=.18\linewidth]{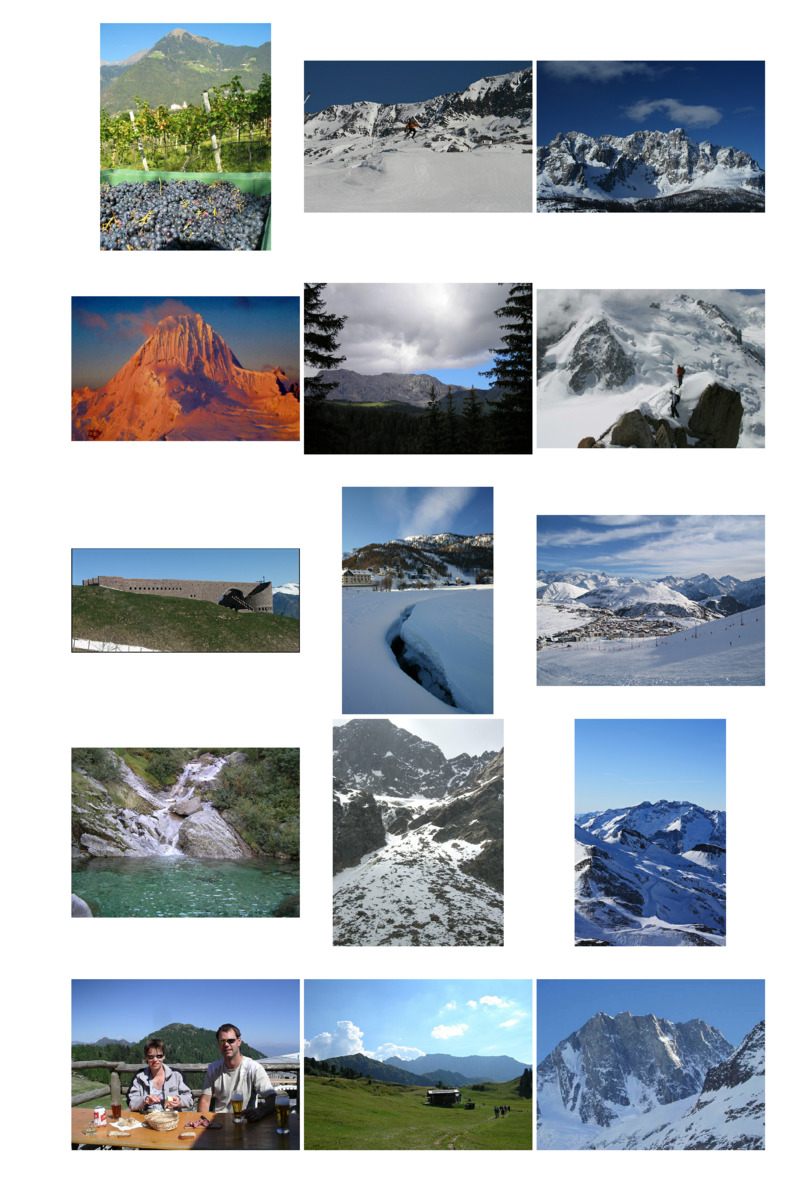}
    \includegraphics[width=.18\linewidth]{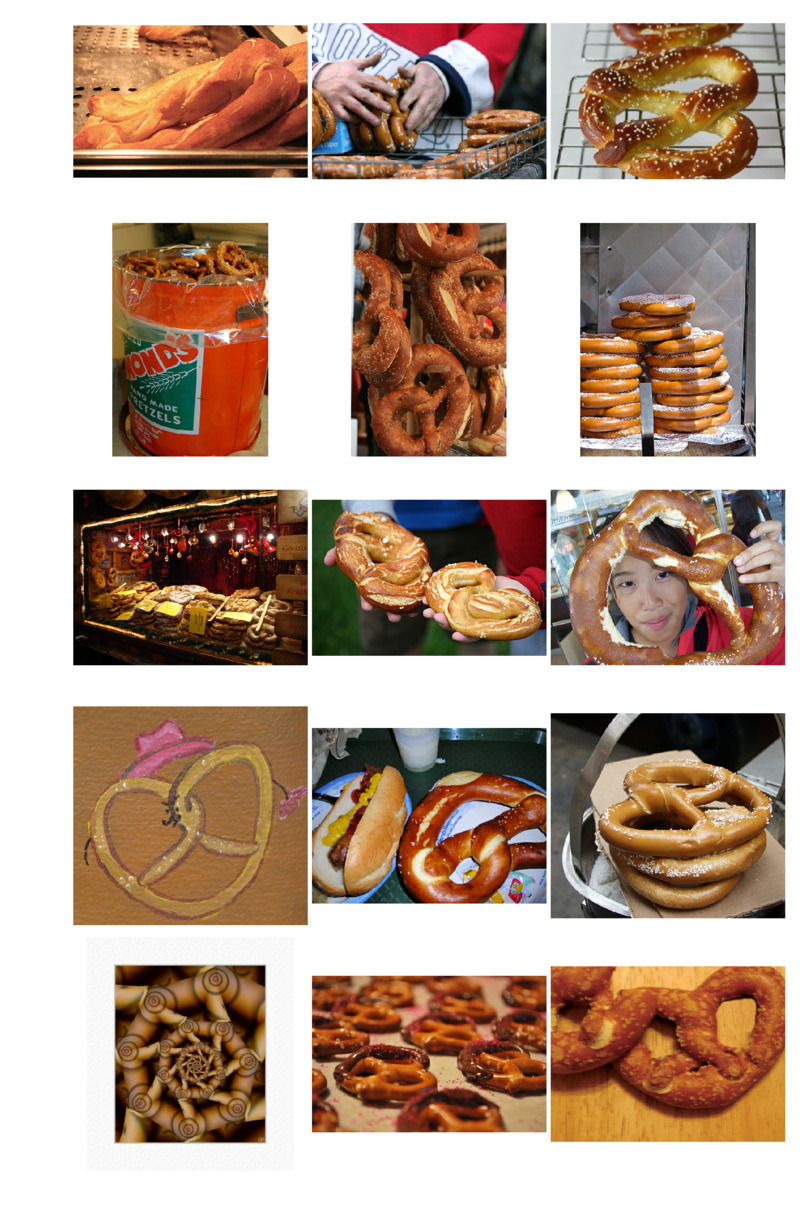}
    \includegraphics[width=.18\linewidth]{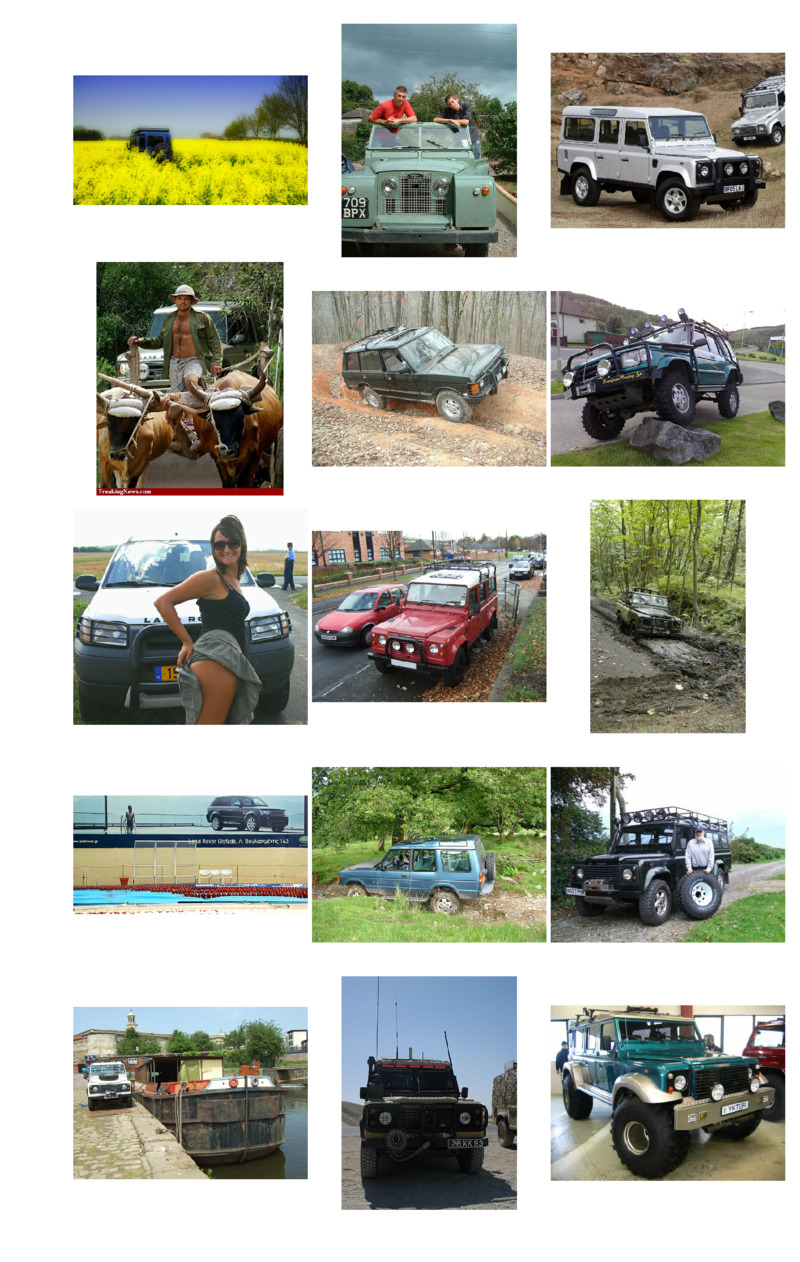}
    \includegraphics[width=.18\linewidth]{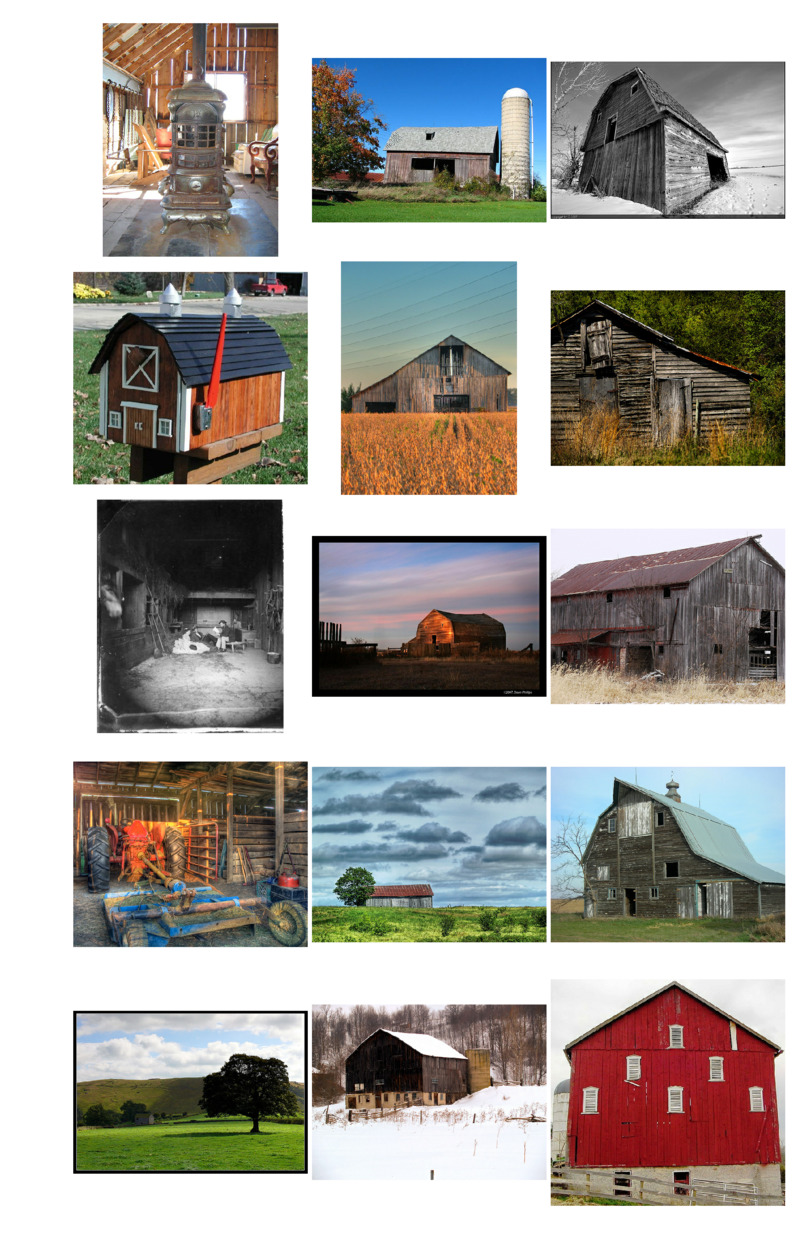}
    \includegraphics[width=.185\linewidth]{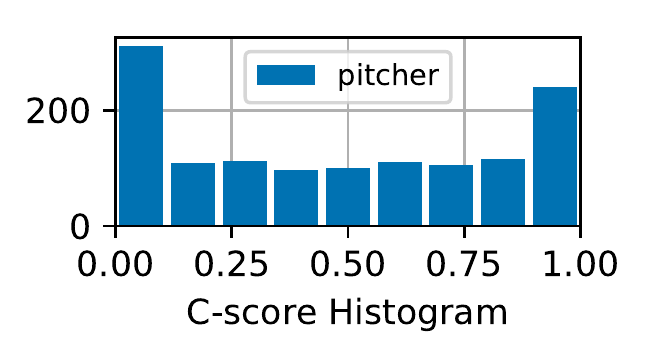}
    \includegraphics[width=.185\linewidth]{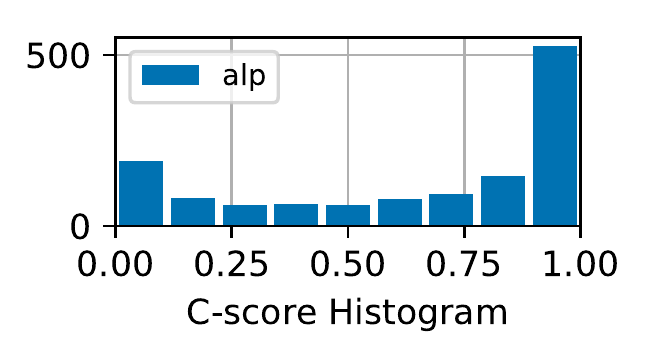}
    \includegraphics[width=.185\linewidth]{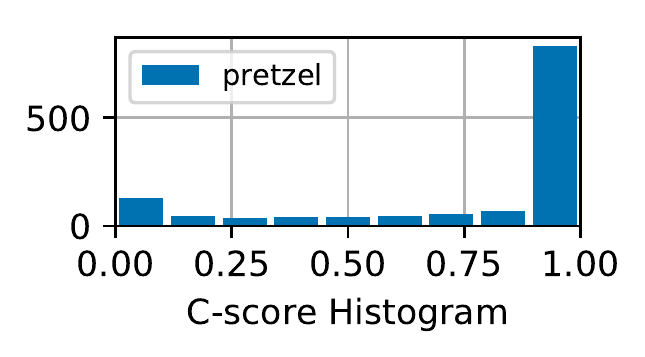}
    \includegraphics[width=.185\linewidth]{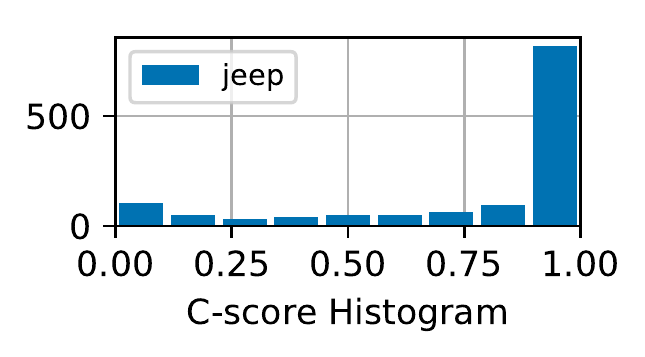}
    \includegraphics[width=.185\linewidth]{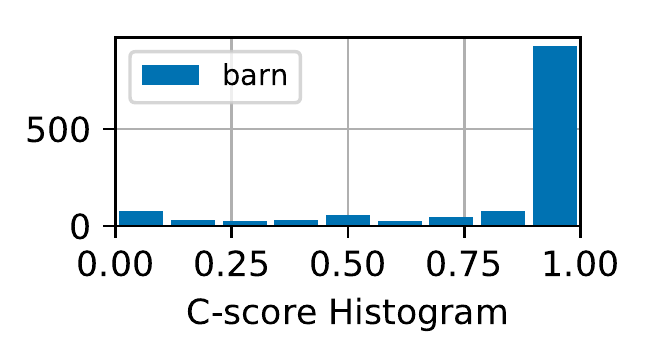}
    \vskip-6pt
    \caption{\small Example images from ImageNet. For each class, the three columns show sampled images from the (\cscore ranked) top 99\%, 35\%, and 1\% percentiles, respectively. The bottom pane shows the histograms of the \cscores in each of the 5 classes.}
    \label{fig:imagenet-per-class-egs-more2}
\end{figure*}

Examples with high, middle and low \cscores from all the 10 classes of MNIST and CIFAR-10 are shown in Figure~\ref{fig:mnist-sweep-full} and Figure~\ref{fig:cifar10-sweep-full}, respectively. The results from the first 60 out of the 100 classes on CIFAR-100 is depicted in Figure~\ref{fig:cifar100-sweep-full}. Figure~\ref{fig:imagenet-per-class-egs-more} and Figure~\ref{fig:imagenet-per-class-egs-more2} show visualizations from ImageNet. Please see the \href{https://pluskid.github.io/structural-regularity/}{project website} for more visualizations.


\section{C-Score Proxies based on Pairwise Distances}

\begin{figure}\centering
    \begin{subfigure}[b]{.48\linewidth}
    \includegraphics[width=\linewidth]{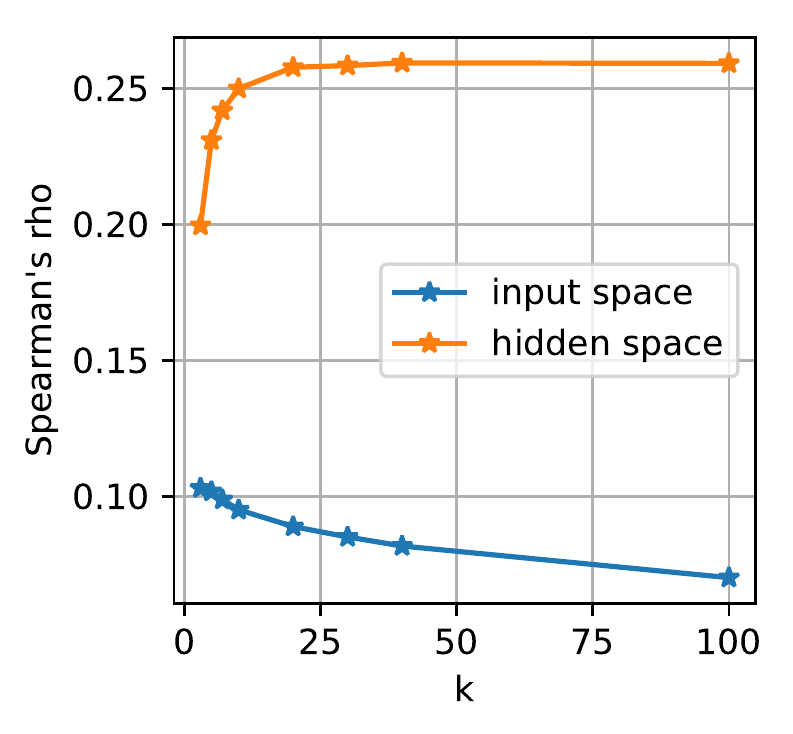}
    \caption{CIFAR-10}
    \end{subfigure}\hfill
    \begin{subfigure}[b]{.48\linewidth}
    \includegraphics[width=\linewidth]{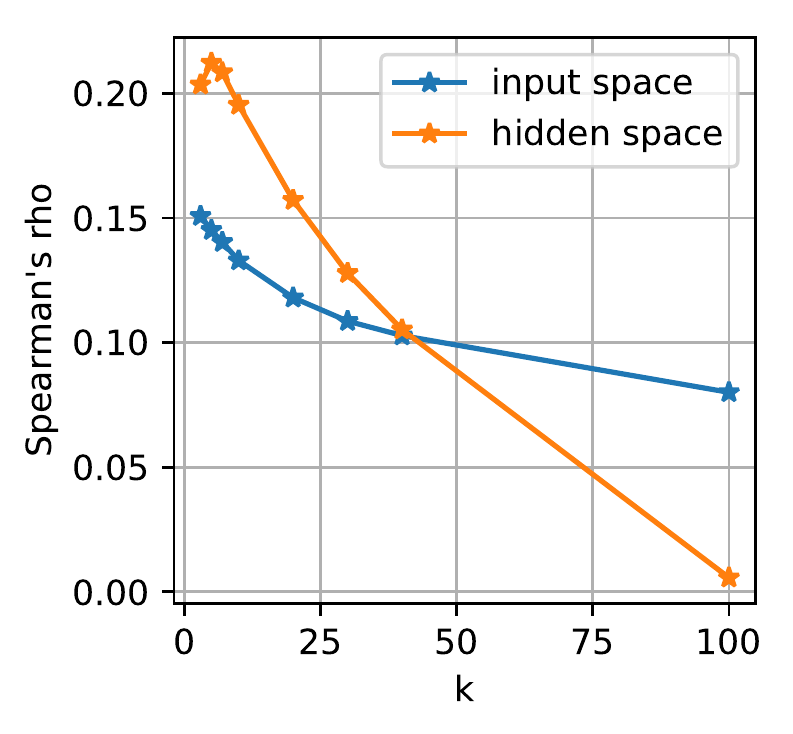}
    \caption{CIFAR-100}
    \end{subfigure}
    \caption{The Spearman's $\rho$ correlation between the \cscore and the score based on LOF with different neighborhood sizes.}
    \label{fig:app-lof-k-tune}
\end{figure}

In the experiments of pairwise distance based \cscore proxies, we use an RBF kernel $K(x,x')=\exp(-\|x-x'\|^2/h^2)$, where the bandwidth parameter $h$ is adaptively chosen as $1/2$ of the mean pairwise Euclidean distance across the data set. For the local outlier factor (LOF) algorithm \citep{Breunig2000}, we use the neighborhood size $k=3$. See Figure~\ref{fig:app-lof-k-tune} for the behavior of LOF across a wide range of neighborhood sizes.

\begin{figure*}
    \includegraphics[width=.09\linewidth]{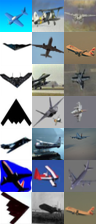}\hspace{1pt}
    \includegraphics[width=.09\linewidth]{figs/kde/viz/cifar10-class_weighted_scores_on_raw-bigpic-class1}\hspace{1pt}
    \includegraphics[width=.09\linewidth]{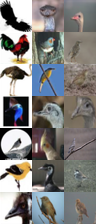}\hspace{1pt}
    \includegraphics[width=.09\linewidth]{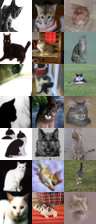}\hspace{1pt}
    \includegraphics[width=.09\linewidth]{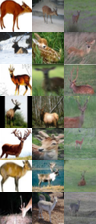}\hfill
    \includegraphics[width=.09\linewidth]{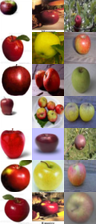}\hspace{1pt}
    \includegraphics[width=.09\linewidth]{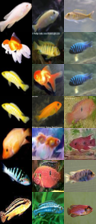}\hspace{1pt}
    \includegraphics[width=.09\linewidth]{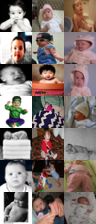}\hspace{1pt}
    \includegraphics[width=.09\linewidth]{figs/kde/viz/cifar100-class_weighted_scores_on_raw-bigpic-class3}\hspace{1pt}
    \includegraphics[width=.09\linewidth]{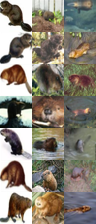}
    \caption{Examples from CIFAR-10 (left 5 blocks) and CIFAR-100 (right 5 blocks). Each block shows a single class; the left, middle, and right columns of a block depict instances with top, intermediate, and bottom ranking according to the relative local density score $\hat{C}^{\pm L}$ in the input space, respectively.}
    \label{fig:cifar-kdeviz-input}
\end{figure*}

\begin{figure*}
    \includegraphics[width=.09\linewidth]{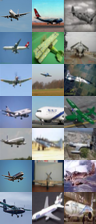}\hspace{1pt}
    \includegraphics[width=.09\linewidth]{figs/kde/viz/cifar10-class_weighted_scores-bigpic-class1}\hspace{1pt}
    \includegraphics[width=.09\linewidth]{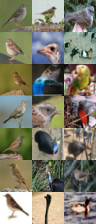}\hspace{1pt}
    \includegraphics[width=.09\linewidth]{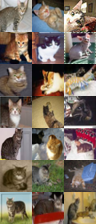}\hspace{1pt}
    \includegraphics[width=.09\linewidth]{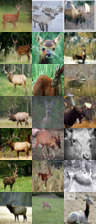}\hfill
    \includegraphics[width=.09\linewidth]{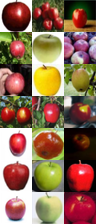}\hspace{1pt}
    \includegraphics[width=.09\linewidth]{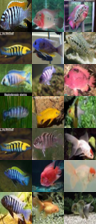}\hspace{1pt}
    \includegraphics[width=.09\linewidth]{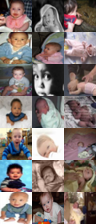}\hspace{1pt}
    \includegraphics[width=.09\linewidth]{figs/kde/viz/cifar100-class_weighted_scores-bigpic-class3}\hspace{1pt}
    \includegraphics[width=.09\linewidth]{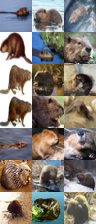}
    \caption{Examples from CIFAR-10  (left 5 blocks) and CIFAR-100 (right 5 blocks). Each block shows a single class; the left, middle, and right columns of a block depict instances with top, intermediate, and bottom ranking according to the relative local density score $\hat{C}^{\pm L}_h$ in the latent representation space of a trained network, respectively.}
    \label{fig:cifar-kdeviz-repr}
\end{figure*}

\subsection{Pairwise Distance Estimation with Gradient Representations}
\label{appendix:gradient}

Most modern neural networks are trained with first order gradient descent based algorithms and variants. In each iteration, the gradient of loss on a mini-batch of training examples evaluated at the current network weights is computed and used to update the current parameter. Let $\nabla_t(\cdot)$ be the function that maps an input-label training pair (the case of mini-batch size one) to the corresponding gradient evaluated at the network weights of the $t$-th iteration. Then this defines a gradient based representation on which we can compute density based ranking scores. The intuition is that in a gradient based learning algorithm, an example is consistent with others if they all compute similar gradients.

Comparing to the hidden representations defined the outputs of a neural network layer, the gradient based representations induce a more natural way of incorporating the label information. In the previous section, we reweight the neighbor examples belonging to a different class by 0 or -1. For gradient based representations, no ad hoc reweighting is needed as the gradient is computed on the loss that has already takes the label into account. Similar inputs with different labels automatically lead to dissimilar gradients. Moreover, this could seamlessly handle labels and losses with rich structures (e.g. image segmentation, machine translation) where an effective reweighting scheme is hard to find.
The gradient based representation is closely related to recent developments on Neural Tagent Kernels (NTK) \citep{jacot2018neural}. It is shown that when the network width goes to infinity, the neural network training dynamics can be effectively approximately via Taylor expansion at the initial network weights. In other words, the algorithm is effectively learning a \emph{linear} model on the \emph{nonlinear} representations defined by $\nabla_0(\cdot)$. This feature map induces the NTK, and connects deep learning to the literature of kernel machines.

Although NTK enjoys nice theoretical properties, it is challenging to perform density estimation on it. Even for the more practical case of \emph{finite width} neural networks, the gradient representations are of extremely high dimensions as modern neural networks general have parameters ranging from millions to billions \citep[e.g.][]{tan2019efficientnet,radford2019language}. As a result, both computation and memory requirements are prohibitive if a naive density estimation is to be computed on the gradient representations. We leave as future work to explore efficient algorithms to practically compute this score.

\section{What Makes an Item Regular or Irregular?}
\label{app:what-makes-item-regular}

\begin{figure}\centering
    \includegraphics[width=\linewidth]{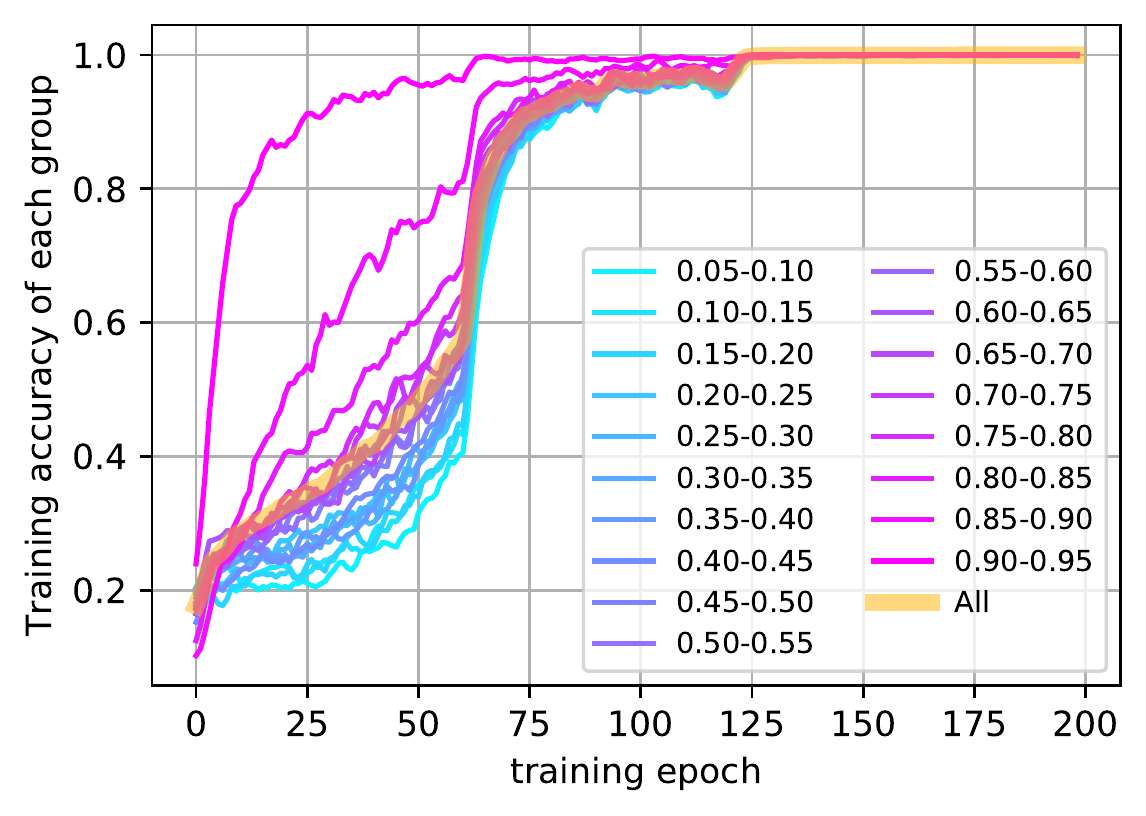}
    \caption{Learning speed of group of examples ranked by \cscores, with equal number (400) of examples in each group via subsampling.}
    \label{fig:learning-speed-equal-group}
\end{figure}

The notion of regularity is primarily coming from the statistical consistency of the example with 
the rest of the population, but less from the intrinsic structure of the example's contents. To illustrate this, we refer back to Figure~4b in the main text, the distribution is uneven between high and low \cscore values. As a result, the high \cscore groups will have more examples than the low \cscore groups. This agrees with the intuition that regularity arises from high probability masses.

To test whether an example with top-ranking \cscore is still highly regular after the density of its neighborhood is reduced, we group the training examples according equal sized bins on the value range of their \cscore values. We then subsample each group to contain an equal number ($\sim 400$) of examples. Then we run training on this new data set and observe the learning speed in each (subsampled) group. The result is shown in Figure~\ref{fig:learning-speed-equal-group}, which is to be compared with the results without group-size-equalizing in Figure~10 in the main text. The following observations can be made:

\begin{enumerate}
    \item The learning curves for many of the groups start to overlap with each other.
    \item The lower ranked groups now learns faster. For example, the lowest ranked group goes above 30\% accuracy near epoch 50. In the run without subsampling (Figure~10a in the main text), this groups is still below 20\% accuracy at epoch 50. The model is now learning with a much smaller data set. Since the lower ranked examples are not highly consistent with the rest of the population, this means there are fewer ``other examples'' to compete with (i.e. those ``other examples'' will move the weights towards a direction that is less preferable for the lower ranked examples). As a result, the lower ranked groups can now learn faster.
    \item On the other hand, the higher ranked groups now learn slower, which is clear from a direct comparison between Figure~10a in the main text and Figure~\ref{fig:learning-speed-equal-group} here. This is because for highly regular examples, reducing the data set size means removing consistent examples --- that is, there are now less ``supporters'' as oppose to less ``competitors'' in the case of lower ranked groups. As a result, the learn speed is now slower.
    \item Even though the learning curves are now overlapping, the highest ranked group and the lowest ranked group are still clearly separated. The potential reason is that while the lower ranked examples can be outliers in many different ways, the highest ranked examples are probably regular in a single (or very few) visual clusters (see the top ranked examples in Figure~\ref{fig:cifar10-sweep-full}). As a result, the within group diversities of the highest ranked groups are still much smaller than the lowest ranked groups.
\end{enumerate}

In summary, the regularity of an example arises from its consistency relation with the rest of the population. A regular example in isolation is no different to an outlier. Moreover, it is also not merely an intrinsic property of the data distribution, but is closely related to the model, loss function and learning algorithms. For example, while a picture with a red lake and a purple forest is likely be considered an outlier in the usual sense, for a model that only uses grayscale information it could be highly regular.

\section{Sensitivity of \cscores to the Number of Models}

\begin{figure}
    \centering
    \includegraphics[width=\linewidth]{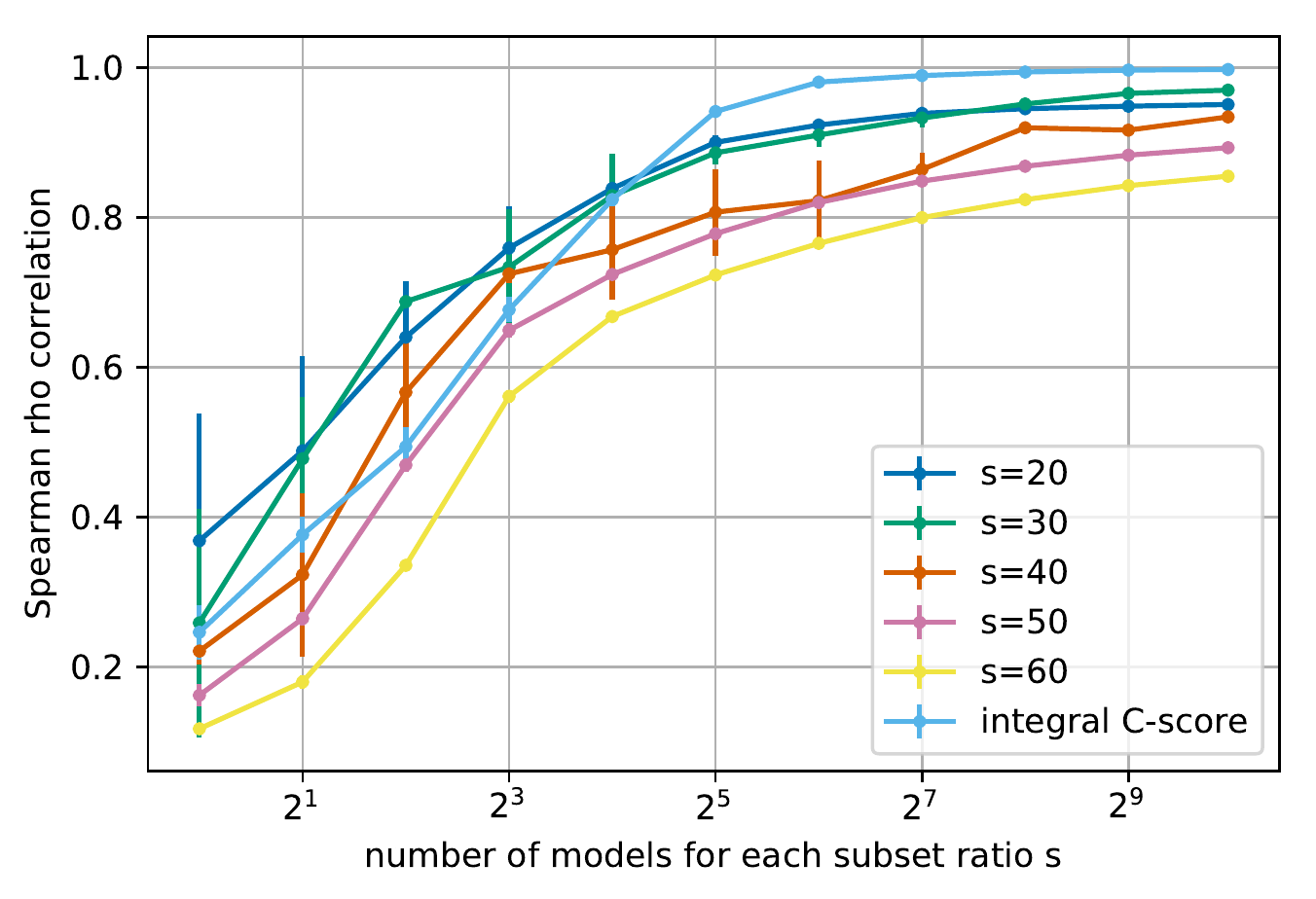}
    \caption{The correlation of \cscores estimated with varying numbers of models (the x-axis) and \cscores estimated with 1,000 independent models. The simulations are run with CIFAR-10, and the error bars show standard deviation from 10 runs.}
    \label{fig:cscore-corr-m}
\end{figure}

We used 2,000 models per subset ratio to evaluate \cscores in our experiments to ensure that we get stable estimates. In this section, we study the sensitivity of \cscores with respect to the number of models and evaluate the possibility to use fewer models in practice. Let $C_{0-2k}$ be the \cscores estimated with the full 2,000 models per subset ratio. We split the 2,000 models for each subset ratio into two halves, and obtain two independent estimates $C_{0-1k}$ and $C_{1k-2k}$. Then for $m\in\{1, 2, 4, 8, 16, 32, 64, 128, 256, 512, 1000\}$, we sample $m$ random models from the first 1,000 split, and estimate \cscores (denoted by $C_m$) based on those models. We compute the Spearman's $\rho$ correlation between each $C_m$ and $C_{1k-2k}$. The results are plotted in Figure~\ref{fig:cscore-corr-m}. The random sampling of $m$ models is repeated 10 times for each $m$ and the error bars show the standard deviations. The figure shows that a good correlation is found for as few as $m=64$ models. However, the integral C-score requires training models for various subset ratios (9 different subset ratios in our simulations), so the total number of models needed is roughly $64 \times 9$. If we want to obtain a reliable estimate of the C-score under a single fixed subset ratio, we find that we need $512$ models in order to get a $> .95$ correlation with $C_{1k-2k}$. So it appears that whether we are computing the integral C-score or the C-score for a particular subset ratio, we need to train on the order of 500-600 models. 

In the analysis above, we have used $C_{1k-2k}$ as the reference scores to compute correlation to ensure no overlapping between the models used to compute different estimates. Note $C_{1k-2k}$ itself is well correlated with the the full estimate from 2,000 models, as demonstrated by the following correlations: $\rho(C_{0-1k},C_{1k-2k})=0.9996$, $\rho(C_{0-1k},C_{0-2k})=0.9999$, and $\rho(C_{1k-2k},C_{0-2k})=0.9999$.


%% file: arxiv.bbl
\begin{thebibliography}{33}
\providecommand{\natexlab}[1]{#1}
\providecommand{\url}[1]{\texttt{#1}}
\expandafter\ifx\csname urlstyle\endcsname\relax
  \providecommand{\doi}[1]{doi: #1}\else
  \providecommand{\doi}{doi: \begingroup \urlstyle{rm}\Url}\fi

\bibitem[Abadi et~al.(2015)Abadi, Agarwal, Barham, Brevdo, Chen, Citro,
  Corrado, Davis, Dean, Devin, Ghemawat, Goodfellow, Harp, Irving, Isard, Jia,
  Jozefowicz, Kaiser, Kudlur, Levenberg, Man\'{e}, Monga, Moore, Murray, Olah,
  Schuster, Shlens, Steiner, Sutskever, Talwar, Tucker, Vanhoucke, Vasudevan,
  Vi\'{e}gas, Vinyals, Warden, Wattenberg, Wicke, Yu, and
  Zheng]{tensorflow2015-whitepaper}
Abadi, M., Agarwal, A., Barham, P., Brevdo, E., Chen, Z., Citro, C., Corrado,
  G.~S., Davis, A., Dean, J., Devin, M., Ghemawat, S., Goodfellow, I., Harp,
  A., Irving, G., Isard, M., Jia, Y., Jozefowicz, R., Kaiser, L., Kudlur, M.,
  Levenberg, J., Man\'{e}, D., Monga, R., Moore, S., Murray, D., Olah, C.,
  Schuster, M., Shlens, J., Steiner, B., Sutskever, I., Talwar, K., Tucker, P.,
  Vanhoucke, V., Vasudevan, V., Vi\'{e}gas, F., Vinyals, O., Warden, P.,
  Wattenberg, M., Wicke, M., Yu, Y., and Zheng, X.
\newblock {TensorFlow}: Large-scale machine learning on heterogeneous systems,
  2015.
\newblock URL \url{https://www.tensorflow.org/}.
\newblock Software available from tensorflow.org.

\bibitem[Bengio et~al.(2009)Bengio, Louradour, Collobert, and
  Weston]{bengio2009}
Bengio, Y., Louradour, J., Collobert, R., and Weston, J.
\newblock Curriculum learning.
\newblock In \emph{Proceedings of the 26th annual international conference on
  machine learning}, pp.\  41--48. ACM, 2009.

\bibitem[Breunig et~al.(2000)Breunig, Kriegel, Ng, and Sander]{Breunig2000}
Breunig, M.~M., Kriegel, H.-P., Ng, R.~T., and Sander, J.
\newblock Lof: identifying density-based local outliers.
\newblock In \emph{Proceedings of the 2000 ACM SIGMOD international conference
  on Management of data}, pp.\  93--104, 2000.

\bibitem[Carlini et~al.(2018)Carlini, Erlingsson, and Papernot]{protypical}
Carlini, N., Erlingsson, U., and Papernot, N.
\newblock Prototypical examples in deep learning: Metrics, characteristics, and
  utility.
\newblock Technical report, OpenReview, 2018.

\bibitem[Caron et~al.(2021)Caron, Touvron, Misra, J{\'e}gou, Mairal,
  Bojanowski, and Joulin]{caron2021emerging}
Caron, M., Touvron, H., Misra, I., J{\'e}gou, H., Mairal, J., Bojanowski, P.,
  and Joulin, A.
\newblock Emerging properties in self-supervised vision transformers.
\newblock \emph{arXiv preprint arXiv:2104.14294}, 2021.

\bibitem[Chen et~al.(2020{\natexlab{a}})Chen, Kornblith, Norouzi, and
  Hinton]{chen2020simple}
Chen, T., Kornblith, S., Norouzi, M., and Hinton, G.
\newblock A simple framework for contrastive learning of visual
  representations.
\newblock In \emph{International conference on machine learning}, pp.\
  1597--1607. PMLR, 2020{\natexlab{a}}.

\bibitem[Chen et~al.(2020{\natexlab{b}})Chen, Kornblith, Swersky, Norouzi, and
  Hinton]{Chen2020-dk}
Chen, T., Kornblith, S., Swersky, K., Norouzi, M., and Hinton, G.
\newblock Big {Self-Supervised} models are strong {Semi-Supervised} learners.
\newblock In \emph{Advances in Neural Information Processing Systems},
  2020{\natexlab{b}}.

\bibitem[Dosovitskiy et~al.(2021)Dosovitskiy, Beyer, Kolesnikov, Weissenborn,
  Zhai, Unterthiner, Dehghani, Minderer, Heigold, Gelly, Uszkoreit, and
  Houlsby]{dosovitskiy2020image}
Dosovitskiy, A., Beyer, L., Kolesnikov, A., Weissenborn, D., Zhai, X.,
  Unterthiner, T., Dehghani, M., Minderer, M., Heigold, G., Gelly, S.,
  Uszkoreit, J., and Houlsby, N.
\newblock An image is worth 16x16 words: Transformers for image recognition at
  scale.
\newblock In \emph{International Conference on Learning Representations}, 2021.

\bibitem[Feldman(2020)]{feldman2019does}
Feldman, V.
\newblock {Does learning require memorization? A short tale about a long tail}.
\newblock In \emph{ACM Symposium on Theory of Computing (STOC)}, 2020.

\bibitem[Feldman \& Zhang(2020)Feldman and Zhang]{fz2019}
Feldman, V. and Zhang, C.
\newblock What neural networks memorize and why: Discovering the long tail via
  influence estimation.
\newblock In \emph{Advances in neural information processing systems}, 2020.

\bibitem[Grill et~al.(2020)Grill, Strub, Altch{\'e}, Tallec, Richemond,
  Buchatskaya, Doersch, Pires, Guo, Azar, Piot, Kavukcuoglu, Munos, and
  Valko]{Grill2020-qw}
Grill, J.-B., Strub, F., Altch{\'e}, F., Tallec, C., Richemond, P.~H.,
  Buchatskaya, E., Doersch, C., Pires, B.~A., Guo, Z.~D., Azar, M.~G., Piot,
  B., Kavukcuoglu, K., Munos, R., and Valko, M.
\newblock Bootstrap your own latent: A new approach to {Self-Supervised}
  learning.
\newblock In \emph{Advances in Neural Information Processing Systems}, 2020.

\bibitem[Hardt et~al.(2016)Hardt, Recht, and Singer]{hardt2016train}
Hardt, M., Recht, B., and Singer, Y.
\newblock Train faster, generalize better: Stability of stochastic gradient
  descent.
\newblock In \emph{International Conference on Machine Learning}, pp.\
  1225--1234. PMLR, 2016.

\bibitem[Jacot et~al.(2018)Jacot, Gabriel, and Hongler]{jacot2018neural}
Jacot, A., Gabriel, F., and Hongler, C.
\newblock Neural tangent kernel: Convergence and generalization in neural
  networks.
\newblock In \emph{Advances in neural information processing systems}, pp.\
  8571--8580, 2018.

\bibitem[Keskar \& Socher(2017)Keskar and Socher]{keskar2017improving}
Keskar, N.~S. and Socher, R.
\newblock Improving generalization performance by switching from adam to sgd.
\newblock \emph{arXiv preprint arXiv:1712.07628}, 2017.

\bibitem[Kingma \& Ba(2015)Kingma and Ba]{adam}
Kingma, D.~P. and Ba, J.
\newblock Adam: A method for stochastic optimization.
\newblock In \emph{International Conference on Learning Representations}, 2015.

\bibitem[Krizhevsky(2009)]{cifar}
Krizhevsky, A.
\newblock Learning multiple layers of features from tiny images.
\newblock Technical Report TR-2009, University of Toronto, 2009.

\bibitem[LeCun et~al.(1998)LeCun, Bottou, Bengio, and
  Haffner]{lecun1998gradient}
LeCun, Y., Bottou, L., Bengio, Y., and Haffner, P.
\newblock Gradient-based learning applied to document recognition.
\newblock \emph{Proceedings of the IEEE}, 86\penalty0 (11):\penalty0
  2278--2324, 1998.

\bibitem[Luo et~al.(2019)Luo, Xiong, Liu, and Sun]{luo2019adaptive}
Luo, L., Xiong, Y., Liu, Y., and Sun, X.
\newblock Adaptive gradient methods with dynamic bound of learning rate.
\newblock In \emph{International Conference on Learning Representations}, 2019.

\bibitem[Mangalam \& Prabhu(2019)Mangalam and Prabhu]{mangalam2019deep}
Mangalam, K. and Prabhu, V.~U.
\newblock Do deep neural networks learn shallow learnable examples first?
\newblock In \emph{ICML 2019 Workshop on Identifying and Understanding Deep
  Learning Phenomena}, 2019.

\bibitem[Melas-Kyriazi(2021)]{melas2021you}
Melas-Kyriazi, L.
\newblock Do you even need attention? a stack of feed-forward layers does
  surprisingly well on imagenet.
\newblock \emph{arXiv preprint arXiv:2105.02723}, 2021.

\bibitem[Netzer et~al.(2011)Netzer, Wang, Coates, Bissacco, Wu, and
  Ng]{netzer2011reading}
Netzer, Y., Wang, T., Coates, A., Bissacco, A., Wu, B., and Ng, A.~Y.
\newblock Reading digits in natural images with unsupervised feature learning.
\newblock In \emph{NIPS Workshop on Deep Learning and Unsupervised Feature
  Learning}, 2011.

\bibitem[Pleiss et~al.(2020)Pleiss, Zhang, Elenberg, and
  Weinberger]{pleiss2019detecting}
Pleiss, G., Zhang, T., Elenberg, E.~R., and Weinberger, K.~Q.
\newblock Detecting noisy training data with loss curves.
\newblock In \emph{International Conference on Learning Representations}, 2020.

\bibitem[Radford et~al.(2019)Radford, Wu, Child, Luan, Amodei, and
  Sutskever]{radford2019language}
Radford, A., Wu, J., Child, R., Luan, D., Amodei, D., and Sutskever, I.
\newblock Language models are unsupervised multitask learners.
\newblock \emph{OpenAI Blog}, 1\penalty0 (8):\penalty0 9, 2019.

\bibitem[Rumelhart \& McClelland(1986)Rumelhart and McClelland]{rumelhart1986}
Rumelhart, D.~E. and McClelland, J.~L.
\newblock \emph{On Learning the Past Tenses of English Verbs}, pp.\  216–271.
\newblock MIT Press, Cambridge, MA, USA, 1986.

\bibitem[Russakovsky et~al.(2015)Russakovsky, Deng, Su, Krause, Satheesh, Ma,
  Huang, Karpathy, Khosla, Bernstein, et~al.]{russakovsky2015imagenet}
Russakovsky, O., Deng, J., Su, H., Krause, J., Satheesh, S., Ma, S., Huang, Z.,
  Karpathy, A., Khosla, A., Bernstein, M., et~al.
\newblock Imagenet large scale visual recognition challenge.
\newblock \emph{International journal of computer vision}, 115\penalty0
  (3):\penalty0 211--252, 2015.

\bibitem[Saxena et~al.(2019)Saxena, Tuzel, and DeCoste]{saxena2019}
Saxena, S., Tuzel, O., and DeCoste, D.
\newblock Data parameters: A new family of parameters for learning a
  differentiable curriculum.
\newblock In Wallach, H., Larochelle, H., Beygelzimer, A., d\textquotesingle
  Alch\'{e}-Buc, F., Fox, E., and Garnett, R. (eds.), \emph{Advances in Neural
  Information Processing Systems 32}, pp.\  11093--11103. Curran Associates,
  Inc., 2019.

\bibitem[Scott et~al.(2018)Scott, Ridgeway, and Mozer]{Scott2018}
Scott, T., Ridgeway, K., and Mozer, M.~C.
\newblock Adapted deep embeddings: A synthesis of methods for k-shot inductive
  transfer learning.
\newblock In Bengio, S., Wallach, H., Larochelle, H., Grauman, K.,
  Cesa-Bianchi, N., and Garnett, R. (eds.), \emph{Advances in Neural
  Information Processing Systems 31}, pp.\  76--85. Curran Associates, Inc.,
  2018.

\bibitem[Tan \& Le(2019)Tan and Le]{tan2019efficientnet}
Tan, M. and Le, Q.~V.
\newblock Efficientnet: Rethinking model scaling for convolutional neural
  networks.
\newblock \emph{arXiv preprint arXiv:1905.11946}, 2019.

\bibitem[Tolstikhin et~al.(2021)Tolstikhin, Houlsby, Kolesnikov, Beyer, Zhai,
  Unterthiner, Yung, Keysers, Uszkoreit, Lucic, et~al.]{tolstikhin2021mlp}
Tolstikhin, I., Houlsby, N., Kolesnikov, A., Beyer, L., Zhai, X., Unterthiner,
  T., Yung, J., Keysers, D., Uszkoreit, J., Lucic, M., et~al.
\newblock Mlp-mixer: An all-mlp architecture for vision.
\newblock \emph{arXiv preprint arXiv:2105.01601}, 2021.

\bibitem[Toneva et~al.(2019)Toneva, Sordoni, Combes, Trischler, Bengio, and
  Gordon]{toneva2018empirical}
Toneva, M., Sordoni, A., Combes, R. T.~d., Trischler, A., Bengio, Y., and
  Gordon, G.~J.
\newblock An empirical study of example forgetting during deep neural network
  learning.
\newblock In \emph{International Conference on Learning Representations}, 2019.

\bibitem[Touvron et~al.(2021)Touvron, Bojanowski, Caron, Cord, El-Nouby, Grave,
  Joulin, Synnaeve, Verbeek, and J{\'e}gou]{touvron2021resmlp}
Touvron, H., Bojanowski, P., Caron, M., Cord, M., El-Nouby, A., Grave, E.,
  Joulin, A., Synnaeve, G., Verbeek, J., and J{\'e}gou, H.
\newblock Resmlp: Feedforward networks for image classification with
  data-efficient training.
\newblock \emph{arXiv preprint arXiv:2105.03404}, 2021.

\bibitem[Wilson et~al.(2017)Wilson, Roelofs, Stern, Srebro, and
  Recht]{wilson2017marginal}
Wilson, A.~C., Roelofs, R., Stern, M., Srebro, N., and Recht, B.
\newblock The marginal value of adaptive gradient methods in machine learning.
\newblock In \emph{Advances in Neural Information Processing Systems}, pp.\
  4148--4158, 2017.

\bibitem[Wu et~al.(2018)Wu, Nagarajan, Kumar, Rennie, Davis, Grauman, and
  Feris]{wu2018blockdrop}
Wu, Z., Nagarajan, T., Kumar, A., Rennie, S., Davis, L.~S., Grauman, K., and
  Feris, R.
\newblock Blockdrop: Dynamic inference paths in residual networks.
\newblock In \emph{Proceedings of the IEEE Conference on Computer Vision and
  Pattern Recognition}, pp.\  8817--8826, 2018.

\end{thebibliography}
